\documentclass[journal]{IEEEtran}
\usepackage{amsmath,graphicx}
\usepackage[T1]{fontenc}
\usepackage{hyperref, times, cite}
\usepackage{url}
\usepackage{enumitem,kantlipsum}
\usepackage{mathtools}
\usepackage{bm, bbm}
\usepackage{cool}
\usepackage{amssymb,amsfonts,amsmath,amsthm,amscd,dsfont,mathrsfs}
\usepackage{graphicx,float,psfrag,epsfig,amssymb}
\usepackage{wrapfig}
\usepackage{relsize}
\usepackage{color}
\usepackage{pict2e}
\usepackage{caption}
\usepackage{nameref}
\usepackage{enumerate}
\usepackage{stackrel}

\usepackage{caption}
\usepackage{subcaption}

\usepackage{amsthm}

\usepackage{cite}
\usepackage[table]{xcolor}
\usepackage{amsmath,amssymb,amsfonts}
\usepackage{graphicx}
\usepackage{textcomp, xcolor}
\usepackage{mathrsfs}
\usepackage{array}
\usepackage{bm}
\usepackage{stfloats}
\usepackage{url}
\usepackage{stackrel}
\usepackage{algorithm}
\usepackage{algcompatible}
\usepackage{algorithmicx}
\usepackage{caption}
\usepackage{subcaption}

\begin{document}
\thispagestyle{empty}

\title{Sparse Graph Learning Under Laplacian-Related Constraints}
\author{ Jitendra K.\ Tugnait 
\thanks{J.K.\ Tugnait is with the Department of 
Electrical \& Computer Engineering,
200 Broun Hall, Auburn University, Auburn, AL 36849, USA. 
Email: tugnajk@auburn.edu . }

\thanks{This work was supported by the National Science Foundation under Grants CCF-1617610 and ECCS-2040536.}}

\maketitle

\renewcommand{\algorithmicrequire}{\textbf{Input:}}
\renewcommand{\algorithmicensure}{\textbf{Output:}}

\begin{abstract}
We consider the problem of learning a sparse undirected graph underlying a given set of multivariate data. We focus on graph Laplacian-related constraints on the sparse precision matrix that encodes conditional dependence between the random variables associated with the graph nodes. Under these constraints the off-diagonal elements of the precision matrix are non-positive (total positivity), and the precision matrix may not be full-rank. We investigate  modifications to widely used penalized log-likelihood approaches to enforce total positivity but not the Laplacian structure. The graph Laplacian can then be extracted from the off-diagonal precision matrix. An alternating direction method of multipliers (ADMM) algorithm is presented and analyzed for constrained optimization under Laplacian-related constraints and lasso as well as adaptive lasso penalties. Numerical results based on synthetic data show that the proposed constrained adaptive lasso approach significantly outperforms existing Laplacian-based approaches. We also evaluate our approach on real financial data. 
\end{abstract}


\newcommand{\algparbox}[1]{\parbox[t]{\dimexpr\linewidth-\algorithmicindent}{#1\strut}}

\section{Introduction} \label{intro}

Graphical models provide a powerful tool for analyzing multivariate data \cite{Lauritzen1996, Dong2019}. In a statistical graphical model, the conditional statistical dependency structure among $p$ random variables $x_1, x_1, \cdots , x_p$, (${\bm x} = [x_1 \; x_2 \; \cdots \; x_p]^\top$), is represented using an undirected graph where there is no edge between nodes $i$ and $j$ iff random variables $x_i$ and $x_j$ associated with these two nodes, are conditionally independent. The precision matrix $\bm{\Omega}$ of ${\bm x} $ encodes this conditional dependence.  Such models for ${\bm x}$ have been extensively studied where a focus has been to estimate $\bm{\Omega}$. Given $n$ samples of ${\bm x}$, in {\it high-dimensional settings}, one estimates $\bm{\Omega}$ under some sparsity constraints; see \cite{Banerjee2008, Danaher2014, Fan2009, Friedman2008, Lam2009, Loh2015, Loh2017, Meinshausen2006, Mohan2014, Ravikumar2011, Rothman2008}.

More recently, several authors have considered Gaussian graphical models under the constraint that the distribution is multivariate totally positive of order 2 ($\mbox{MTP}_2$), or equivalently, that all partial correlations are non-negative (see \cite{Lauritzen2019,  Wang2020} and references therein). Such models are also known as attractive Gaussian random fields \cite{Slawski2015}. Note that a Gaussian distribution is $\mbox{MTP}_2$ if and only if its precision matrix $\bm{\Omega}$ is an M-matrix, i.e., ${\Omega}_{ij} \le 0$ for all $i \ne j$ \cite{Karlin1983}. As discussed in \cite{Wang2020}, $\mbox{MTP}_2$ is a strong form of positive dependence, which is relevant for modeling in various applications. A large majority of the prior work does not impose total positivity.
  
Graphical models have also been inferred from consideration other than statistical \cite{Dong2019}. One class of graphical models are based on signal smoothness \cite{Dong2016, Dong2019, Kalofolias2016, Kalofolias2019} where graph learning from data becomes equivalent to estimation of the graph Laplacian matrix \cite{Kalofolias2016, Dong2019}. The graph Laplacian is positive semi-definite with non-positive off-diagonal entries, hence, can be viewed as rank-deficient precision matrix for an $\mbox{MTP}_2$ Gaussian random vector. Another set of approaches are based on statistical considerations under the graph Laplacian constraint \cite{Pavez2016, Pavez2018, Dong2019, Egilmez2017, Kumar2020, Ying2020} where Laplacian ${\bm L}$ (or a generalized version)  plays the role of the precision matrix ${\bm \Omega}$. Thus, under Gaussian distribution we have an $\mbox{MTP}_2$ model. A key contribution of \cite{Ying2020} has been to show that under convex lasso ($\ell_1$) penalty, Laplacian-constrained log-likelihood approaches do not yield sparse graphs; non-convex penalties are required; see also \cite{Zhang2020}.

Graph Laplacian matrix has been extensively used for embedding, manifold learning, clustering and semi-supervised learning \cite{Belkin2001, Belkin2006, Bengio2006, Luxburg2007, Zhu2003, Zhou2011}; see \cite{Kalofolias2016, Dong2019} for further references to applications to web page categorization with graph information, etc., and \cite{Fracastoro2020} for graph-based transform coding where learning of the graph Laplacian plays a key role.

Recent reviews of various graph learning approaches may be found in \cite{Qiao2018} and \cite{Xia2021}. A large variety of graph learning models and approaches exist, motivated by diverse applications in signal processing, machine learning, and other areas. In \cite{Xia2021} existing graph learning methods are classified into four broad categories: graph signal processing based methods, matrix factorization based methods, random walk based methods, and deep learning based methods. In terms of these four categories, our approach falls in the category of graph signal processing based methods. On the other hand, \cite{Qiao2018} categorizes graph learning methods based on two graph construction steps: (1) determine the edge set ${\cal E}$ (see Sec.\ \ref{PFRW1}), called $E$-step, and (2) based on ${\cal E}$, determine the edge weight matrix ${\bm W}$ (see Sec.\ \ref{PFRW1}), called $W$-step, even though in some methods these two steps may be merged into one, or the second step may be executed first yielding ${\bm W}$ which then determines ${\cal E}$. For instance, our approach yields a matrix equivalent to ${\bm W}$ which then determines ${\cal E}$ (see Sec.\ \ref{PFRW1}). 

A class of graph learning approaches are motivated by specific application tasks such as clustering and semi-supervised classification. Examples of such approaches include \cite{Kang2020, Kang2021} and relevant references in  \cite{Qiao2018} and \cite{Xia2021}. In such approaches an important consideration is how to incorporate prior information relevant to the intended application, in the graph model. For instance, both local and global structure information is incorporated in the model of \cite{Kang2021}, together with a rank constraint on the graph Laplacian to reflect the number of clusters. As noted in \cite{Qiao2018}, ``... how to select a suitable graph construction/learning strategy in practice ... is a challenging problem without a universal solution, since it depends on many factors ...'' 

\subsection{Our Contributions} We investigate  modifications to widely used penalized log-likelihood approaches to enforce total positivity but not the Laplacian structure. The graph Laplacian can then be extracted from the off-diagonal precision matrix. We use log-sum penalty \cite{Candes2008} resulting in adaptive lasso (initialized with lasso), and our approach does not require prior knowledge of the nature of the graph Laplacian (how many components, generalized or not, etc.). An alternating direction method of multipliers (ADMM) algorithm is presented for constrained optimization under total positivity. Numerical results based on synthetic data show that the proposed constrained adaptive lasso approach significantly outperforms existing Laplacian-based approaches \cite{Kalofolias2016, Egilmez2017, Ying2020}.

\subsection{Outline and Notation} 
The rest of the paper is organized as follows. In Sec.\ \ref{PFRW} we formulate the problem for the case where precision matrix ${\bm \Omega}$ is full rank. Past related work for the two cases, ${\bm \Omega}$ is full rank and ${\bm \Omega}$ is rank-deficient, is discussed in Sec.\ \ref{PFRW}. An ADMM algorithm is presented in Sec.\ \ref{sADMM} to optimize the proposed cost function and a pseudocode for the ADMM algorithm is given in Algorithm \ref{alg0}. In Sec.\ \ref{ThA} we analyze consistency (Theorem 1) and sparsistency (Theorem 2) of the proposed approach. Numerical results based on synthetic as well as real data are presented in Sec.\ \ref{SE} to illustrate the proposed approach.  Proofs of Theorems 1 and 2 are given in the two appendices.

We use ${\bm S} \succeq 0$ and ${\bm S} \succ 0$ to denote that the symmetric matrix ${\bm S}$ is positive semi-definite and positive definite, respectively. The set of real numbers is denoted by $\mathbb{R}$. For a set $V$, $|V|$ or $\mbox{card}(V)$ denotes its cardinality, i.e., the number of elements in $V$. Given ${\bm A} \in \mathbb{R}^{p \times p}$, we use $\phi_{\min }({\bm A})$, $\phi_{\max }({\bm A})$, $|{\bm A}|$ and $\mbox{tr}({\bm A})$ to denote the minimum eigenvalue, maximum eigenvalue, determinant and  trace of ${\bm A}$, respectively, {and ${\bm A}^\dagger$ to denote its pseudo-inverse}. For ${\bm B} \in \mathbb{R}^{p \times q}$, we define its operator norm, the Frobenius norm and the vectorized $\ell_1$ norm, respectively, as $\|{\bm B}\| = \sqrt{\phi_{\max }({\bm B}^\top  {\bm B})}$, $\|{\bm B}\|_F = \sqrt{\mbox{tr}({\bm B}^\top  {\bm B})}$ and $\|{\bm B}\|_1 = \sum_{i,j} |B_{ij}|$, where $B_{ij}$ is the $(i,j)$-th element of ${\bm B}$ (also denoted by $[{\bm B}]_{ij}$). Given ${\bm A} \in \mathbb{R}^{p \times p}$, ${\bm A}^+ = \mbox{diag}({\bm A})$ is a diagonal matrix with the same diagonal as ${\bm A}$, and  ${\bm A}^- = {\bm A} - {\bm A}^+$ is ${\bm A}$ with all its diagonal elements set to zero. The symbol $\otimes$ denotes the matrix Kronecker product and ${\bm 1}_A$ denotes the indicator function (which equals 1 if $A$ is true, else 0). For ${\bm y}_n, {\bm x}_n \in \mathbb{R}^p$, ${\bm y}_n \asymp {\bm x}_n$ means that ${\bm y}_n = {\cal O}({\bm x}_n)$ and ${\bm x}_n = {\cal O}({\bm y}_n)$, where the latter means there exists $0 < M < \infty$ such that $\|{\bm x}_n\| \le M \|{\bm y}_n\|$ $\forall n \ge 1$.  The notation ${\bm y}_n = {\cal O}_P({\bm x}_n)$ for random vectors ${\bm y}_n, {\bm x}_n \in \mathbb{R}^p$ means that for any $\varepsilon > 0$, there exists $0 < M < \infty$ such that $P ( \|{\bm y}_n\| \le M \|{\bm x}_n\|) \ge 1 - \varepsilon$ $\forall n \ge 1$.

\section{Problem Formulation and Related Work} \label{PFRW}
In this section we formulate the problem for the case where precision matrix ${\bm \Omega}$ is full rank, but later in simulations, we apply it to rank-deficient ${\bm \Omega}$ also. Past related work for the two cases, ${\bm \Omega}$ is full rank and ${\bm \Omega}$ is rank-deficient, is also discussed.

\subsection{Graphical Models and Graph Laplacians} \label{PFRW1}
An undirected simple weighted graph is denoted ${\cal G} = \left( V, {\cal E}, {\bm W} \right)$ where $V = \{1,2, \cdots , p\} =[p]$ is the set of $p$ nodes, ${\cal E} \subseteq [p] \times [p]$ is the set of undirected edges, and ${\bm W} = {\bm W}^\top \in {\mathbb R}^{p \times p}$ stores the non-negative weights $W_{ij} \ge 0$ associated with the undirected edges. If $W_{ij} > 0$, then edge $\{i,j\} \in {\cal E}$, otherwise edge $\{i,j\} \not\in {\cal E}$. In a simple graph there are no self-loops or multiple edges, so ${\cal E}$ consists of distinct pairs $\{i,j\}$, $i \ne j$ and $W_{ii}=0$. In graphical models of data variables ${\bm x} \in \mathbb{R}^p$, a weighted graph ${\cal G} = \left( V, {\cal E}, {\bm W} \right)$ (or unweighted ${\cal G} = \left( V, {\cal E} \right)$) with $|V|=p$ is used to capture relationships between the $p$ variables $x_i$'s \cite{Lauritzen1996, Dong2019}. If $\{i,j\} \in {\cal E}$, then $x_i$ and $x_j$ are related in some sense, with higher $W_{ij}$ indicating stronger similarity or dependence. A statistical graphical model ${\cal G}$ is a conditional independence graph (CIG) where $\{i,j\} \not\in {\cal E}$ iff $x_i$ and $x_j$ are conditionally independent. In particular, Gaussian graphical models (GGMs) are CIGs where ${\bm x}$ is multivariate Gaussian. Suppose ${\bm x}$ has positive semi-definite covariance matrix $\bm{\Sigma}$ with  precision matrix $\bm{\Omega} = \bm{\Sigma}^\dagger$. Then ${\Omega}_{ij}$, the $(i,j)$-th element of  $\bm{\Omega}$, is zero iff $x_i$ and $x_j$ are conditionally independent.

The (combinatorial) graph Laplacian of ${\cal G} = \left( V, {\cal E}, {\bm W} \right)$ is defined as ${\bm L}= {\bm D}-{\bm W}$ where ${\bm D}$ is the diagonal weighted degree matrix with $D_{ii} = \sum_{j=1}^p W_{ij}$. This makes rank(${\bm L}) < p$ and off-diagonal elements $L_{ij}=-W_{ij} \le 0$ for $i \ne j$. A generalized graph Laplacian is defined as ${\bm L}_g = {\bm L} + {\bm V}$ where ${\bm V}$ is diagonal \cite{Egilmez2017}. If all diagonal elements are ${\bm V}$ are strictly positive, then ${\bm L}_g$ is positive-definite. There has been considerable recent interest in GGMs where one takes ${\bm \Omega}={\bm L}$ (\cite{Egilmez2017, Kumar2020, Ying2020}), or ${\bm \Omega}={\bm L}_g \succ {\bm 0}$ (\cite{Lake2010, Slawski2015, Pavez2016, Pavez2018, Egilmez2017, Soloff2020, Wang2020}). Both cases result in $\Omega_{ij} \le 0$ for $i \ne j$, and this model is addressed in this paper. Our objective is determine ${\cal E}$ and ${\bm L}= {\bm D}-{\bm W}$ for both cases. We estimate ${\bm \Omega}$ as $\hat{\bm \Omega}$ under the constraint $\Omega_{ij} \le 0$ for $i \ne j$, and then set $\hat{\bm W} = - {\bm \Omega}^-$ and $\hat{\bm L}= \hat{\bm D}-\hat{\bm W}$.

\subsection{Full-Rank Precision Matrix Under Total Positivity} \label{sFRPM}
Suppose we are given $n$ i.i.d.\ observations $\{{\bm x}(t)\}_{t=1}^{n}$, ${\bm x} \in \mathbb{R}^{p}$, where ${\bm x}$ is zero-mean Gaussian with covariance ${\bm \Sigma}$ and precision matrix ${\bm \Omega}$. 
In graphical lasso \cite{Friedman2008}, with $\hat{\bm \Sigma} = \frac{1}{n} \sum_{t=1}^n {\bm x}(t) {\bm x}^\top (t)$), one seeks ${\bm \Omega}$ to yield $\min_{{\bm \Omega} \succ {\bm 0} } f_L({\bm \Omega})$ where 
\begin{equation} 
   f_L({\bm \Omega}) = \mbox{tr}({\bm \Omega} \hat{\bm \Sigma} )  
	                     - \ln (|{\bm \Omega}|) + \lambda \|{\bm \Omega}^-\|_1 \, ,
	\label{aneq200}
\end{equation}
$\lambda \|{\bm \Omega}^-\|_1 = \lambda \sum_{i \ne j} |{\Omega}_{ij}|$ is the lasso penalty with $\lambda >0$. In this paper we investigate approaches for the case where we have an additional constraint ${\bm \Omega} \in {\cal V}_p$ where ${\cal V}_p$ is the space of of all $p \times p$ matrices ${\bm V}$ that are symmetric with non-positive off-diagonal elements
\begin{align} 
   {\cal V}_p = & \Big\{ {\bm V} \in \mathbb{R}^{p \times p} \, : \, {\bm V} = {\bm V}^\top , \; V_{ij} \le 0, 
	\; i\ne j \Big\} \, . \label{neq130}
\end{align}
The convex penalty $\lambda \|{\bm \Omega}^-\|_1$ in (\ref{aneq200}) is replaced with the nonconvex log-sum penalty (LSP) $\sum_{i\ne j} P_{\lambda}(\Omega_{ij})$ motivated by \cite{Candes2008} (and \cite{Zou2006}), defined as ($\epsilon$ is small)
\begin{align} 
   P_{\lambda}(\theta) = \lambda \ln \left(1+ |\theta|/\epsilon \right) \, , \; \epsilon > 0 \, .
	\label{neq130a}
\end{align}
That is, replace (\ref{aneq200}) with (\ref{aneq200a})
\begin{equation} 
   f_{LSP}({\bm \Omega}) = \mbox{tr}({\bm \Omega} \hat{\bm \Sigma} )  
	                     - \ln (|{\bm \Omega}|) + \lambda \sum_{i \ne j} \ln \left(1+ |\Omega_{ij}|/\epsilon \right) \, ,
	\label{aneq200a}
\end{equation}
and seek solution to 
\begin{equation} 
   \min_{{\bm \Omega} \succ {\bm 0}, \; {\bm \Omega} \in {\cal V}_p } \, f_{LSP}({\bm \Omega}) \, .
	\label{aneq200d}
\end{equation} 

As for the SCAD (smoothly clipped absolute deviation) penalty in \cite{Lam2009}, we solve the nonconvex problem (\ref{aneq200d}) iteratively, where in each iteration, the problem is convex. Using $\partial P_{\lambda}(|\theta|) / \partial |\theta| = \lambda /(|\theta| + \epsilon)$, a local linear approximation to $P_{\lambda}(|\theta|)$ around $\theta_0$ yields a symmetric linear function
\begin{equation}
  P_{\lambda}(|\theta|) \approx P_{\lambda}(|\theta_0|) 
	 + \frac{\lambda}{|\theta_0| + \epsilon} (|\theta| - |\theta_0|) \, .
\end{equation}
With $\theta_0$ fixed, we need to consider only the term dependent upon $\theta$ for optimization w.r.t.\ $\theta$:
\begin{equation}  \label{local}
  P_{\lambda}(|\theta|) 
	 \, \rightarrow \, \frac{\lambda}{|\theta_0| + \epsilon} |\theta| \, .
\end{equation}
Suppose we have a ``good'' initial solution $\bar{\bm \Omega}$ to the problem (from e.g., using lasso $f_{L}({\bm \Omega})$ instead of $f_{LSP}({\bm \Omega})$). Then, given $\bar{\bm \Omega}$, using the local linear approximation to $P_{\lambda}(\Omega_{ij})$ as in (\ref{local}), after ignoring terms dependent upon $\bar{\bm \Omega}$, we have
\begin{align} 
   f_{LSP}({\bm \Omega}) \propto & \; \mbox{tr}({\bm \Omega} \hat{\bm \Sigma} )   - \ln (|{\bm \Omega}|) 
	  +  \sum_{i \ne j} \lambda_{ij} |{\Omega}_{ij}|  , \label{neq2000} \\
		\lambda_{ij} = & \frac{\lambda}{|\bar{\Omega}_{ij}| + \epsilon} \, .
	\label{neq2000a}
\end{align}
Therefore, in the next iteration we seek
\begin{equation}
  \hat{\bm \Omega} \, = \,  
   \min_{{\bm \Omega} \succ {\bm 0}, \; {\bm \Omega} \in {\cal V}_p } 
	\Big\{ \mbox{tr}({\bm \Omega} \hat{\bm \Sigma} )   - \ln (|{\bm \Omega}|) 
	+  \sum_{i \ne j}  \lambda_{ij} |{\Omega}_{ij}| \Big\} 
	\label{neq2010}
\end{equation}
with $\lambda_{ij}$ as in (\ref{neq2000a}).
This is then adaptive lasso \cite{Zou2006}; strictly speaking, \cite{Zou2006} has $\epsilon = 0$. If we initialize with $\bar{\Omega}_{ij} = 0$ (or some other constant) for all $i \ne j$, we obtain a lasso cost.

Since in each iteration we have a convex optimization problem, we obtain a global minimum to the linearized problem. But since the original problem (\ref{aneq200d}) is nonconvex because LSP is nonconvex, overall, we are only guaranteed a local minimum of the original problem. The unconstrained lasso minimizer of $f_L({\bm \Omega})$ specified in (\ref{aneq200}) is consistent \cite{Rothman2008} (where local consistency implies global consistency), and consistency of the constrained lasso minimizer of $f_L({\bm \Omega})$ under the additional constraint  ${\bm \Omega} \in {\cal V}_p$ follows as in the proof of Theorem 1 in Sec.\ \ref{ThA}. Therefore, if we initialize with the constrained lasso minimizer of $f_L({\bm \Omega})$, i.e., choose $\bar{\Omega}_{ij}$ to be constrained lasso minimizer, we should expect the local minimum of the iterative solution to the original nonconvex problem to be close to the global minimum.

\subsection{Related work and comparisons} There are two lines of related work on statistical models: one dealing with $\mbox{MTP}_2$ models where the precision matrix is full-rank \cite{Slawski2015, Soloff2020, Wang2020}, and the other dealing with explicit Laplacian constraint \cite{Dong2019, Egilmez2017, Kumar2020, Ying2020, Zhang2020}. Note that neither sparsity nor large sample size is required under $\mbox{MTP}_2$ assumption for existence of precision matrix estimate \cite{Slawski2015}.  Full-rank precision matrix assumption is central to \cite{Slawski2015, Soloff2020, Wang2020}, whereas empirical evidence suggests that our approach does not require it. \cite{Soloff2020} deals with certain theoretical guarantees under Stein loss for precision matrix estimation. \cite{Wang2020} does not estimate the precision matrix, only the edges. \cite{Kumar2020, Ying2020, Zhang2020} assume single-component Laplacians (only one zero eigenvalue) as precision matrix whereas in this paper it is not required. Our approach can handle multi-component Laplacians. \cite{Kumar2020} uses some spectral constraint which we do not, while  \cite{Ying2020, Zhang2020}  use non-convex penalties (SCAD and related minimax concave penalty (MCP) in \cite{Ying2020}, and MCP in \cite{Zhang2020}) while we use non-convex LSP. In our approach, replacing LSP with SCAD yielded only marginal improvements over the convex lasso constraint in our numerical results. \cite{Dong2019, Egilmez2017} use convex $\ell_1$-penalty, and as shown in \cite{Ying2020, Zhang2020}, under $\ell_1$-penalty, Laplacian-constrained log-likelihood approaches do not yield sparse graphs. A signal smoothness based approach is used in \cite{Kalofolias2016} who directly estimates the weight matrix ${\bm W}$ and then constructs ${\bm L}$ from it. We estimate ${\bm \Omega}$, then set $\hat{\bf W} = - {\bf \Omega}^-$, and then follow \cite{Kalofolias2016} in setting $\hat{\bm L}= \hat{\bm D}-\hat{\bm W}$.

\section{ADMM Solution} \label{sADMM}
 To solve (\ref{neq2010}) we will use alternating direction method of multipliers (ADMM) \cite{Boyd2010} after variable splitting. Using variable splitting, consider
\begin{align} 
 \min_{\bm{\Omega} \succ {\bm 0}, \; {\bm V} \in {\cal V}_p } & \Big\{  \mbox{tr} ( \hat{\bm{\Sigma}} \bm{\Omega}  ) 
          -   \ln(|\bm{\Omega}|) + \sum_{i \ne j}  \lambda_{ij} |V_{ij}|   \Big\}  \label{eqth2_4000} \\
					 \mbox{ subject to  } & \bm{\Omega} = {\bm V}  \, .    \label{eqth2_4000a}   
\end{align}
The scaled augmented Lagrangian for this problem is \cite{Boyd2010}
\begin{align} 
 L_\rho = \mbox{tr} ( \hat{\bm{\Sigma}} \bm{\Omega}  ) 
          -   \ln(|\bm{\Omega}|) + \sum_{i \ne j}  \lambda_{ij} |V_{ij}|   
					 + \frac{\rho}{2}  \| {\bm V} - \bm{\Omega}  + {\bm U}\|^2_F   \label{eqth2_4010}  
\end{align}
where ${\bm U}$ is the dual variable, and $\rho >0$ is the penalty parameter. Given the results $ \bm{\Omega}^{(k)}, {\bm V}^{(k)}, {\bm U}^{(k)}$ of the $k$th iteration, in the $(k+1)$st iteration, an ADMM algorithm executes three updates:
\begin{itemize}
\item[(a)] $\bm{\Omega}^{(k+1)} \leftarrow \arg \min_{\bm{\Omega}} \, L_a(\bm{\Omega}) ,\;\;
           L_a(\bm{\Omega}) := \mbox{tr} ( \hat{\bm{\Sigma}} \bm{\Omega}  ) 
          -   \ln(|\bm{\Omega}|) + \frac{\rho}{2}  \| {\bm V}^{(k)} - \bm{\Omega} + {\bm U}^{(k)}\|^2_F $
\item[(b)] ${\bm V}^{(k+1)}  \leftarrow \arg \min _{ {\bm V} \in {\cal V}_p } L_b({\bm V}) , \;\;
          L_b({\bm V}) :=   \sum_{i \ne j}  \lambda_{ij} |V_{ij}|  
					+ \frac{\rho}{2}  \| {\bm V} - \bm{\Omega}^{(k+1)}  + {\bm U}^{(k)} \|^2_F$
\item[(c)] ${\bm U}^{(k+1)} \leftarrow {\bm U}^{(k)}  +
   \left( {\bm V}^{(k+1)} - \bm{\Omega}^{(k+1)}  \right)$
\end{itemize}
Solution to update (a) follows from \cite[Sec.\ 6.5]{Boyd2010} and is given in step 5 of Algorithm \ref{alg0}, presented later in this section.  For any $\rho > 0$, by construction, $\bm{\Omega}^{(k+1)} \succ {\bm 0}$. This is so even if we apply the algorithm to a problem with true ${\bm \Omega} \succeq {\bm 0}$.

In update (b) notice that $L_b({\bm V})$ is completely separable w.r.t.\ each element $V_{ij}$. Therefore, we solve $V_{ij}^{(k+1)}  \leftarrow \arg \min_{{\textstyle\mathstrut} V_{ij} \le 0, \; i\ne j} J_{ij}(V_{ij})$, where 
$J_{ij}(V_{ij})  :=   \lambda_{ij} |V_{ij}| {\bm 1}_{i \neq  j}
					+ \frac{\rho}{2}  (  V_{ij} - [\bm{\Omega}^{(k+1)}  - {\bm U}^{(k)} ]_{ij} )^2$. 
We claim that the  solution is given by
\begin{align}  \label{soln}
    V_{ij}^{(k+1)} & = 
	 \left\{ \begin{array}{ll}
			    [ \bm{\Omega}^{(k+1)} - {\bm U}^{(k)} ]_{ii} & \mbox{  if } i=j\\
					S_{neg}([\bm{\Omega}^{(k+1)} - {\bm U}^{(k)} ]_{ij}, \frac{ \lambda_{ij}}{\rho}) 
					   &  \mbox{ if } i \neq j \end{array} \right.
\end{align}
where, with $(a)_+ := \max(0,a)$ and $(a)_- := \min(0,a)$,
\[
   S_{neg}(a, \beta) := (1-\beta/|a|)_+ a_-
\]
denotes scalar soft thresholding for negative values of $a$ and hard thresholding for $a > 0$. When $i=j$, we need to minimize only $(  V_{ij} - [\bm{\Omega}^{(k+1)}  - {\bm U}^{(k)} ]_{ij} )^2$ w.r.t.\ $V_{ii}=V_{ij}$, thus the given solution follows. For constrained optimization under $V_{ij} \le 0$, after setting $A_{ij} = [\bm{\Omega}^{(k+1)}  - {\bm U}^{(k)} ]_{ij}$,  consider the Lagrangian $L_v$
\begin{equation}
  L_v = \lambda |V_{ij}|  + \frac{\rho}{2}  (  V_{ij} - A_{ij} )^2 + \nu V_{ij}
\end{equation}
where $\nu \ge 0$ is the Lagrange multiplier for the inequality constraint $V_{ij} \le 0$. With $v^*$ denoting an optimal solution, the KKT conditions for minimization are
\begin{align}  
    0 \in \partial L_v & = \lambda t + \rho (v^* - A_{ij}) + \nu  \\
		      \nu v^* = 0 \\
					\nu \ge 0 \\
					v^* \le 0 
\end{align}
where $\partial L_v$ denotes the subdifferential of $L_v$ at $v^*$ and  
\begin{equation}
   t =  \left\{ \begin{array}{ll}
			   v^* / |v^*| & \mbox{if } v^* \neq 0 \\
					\in \{ u \,:\, |u| \le 1, \; u \in \mathbb{R} \} & \mbox{if } v^* = 0 \end{array} \right. 
\end{equation}
When $A_{ij} > 0$, our claimed solution is $v^* = 0$. We need to check if $\nu \ge 0$ and $0 \in \partial L_v$ for some $|t| \le 1$. The choice $t=0$ and $\nu = \rho A_{ij} >0$ satisfies the KKT conditions. When $A_{ij} \le 0$, our claimed solution is the well-known soft-thresholding solution which satisfies the KKT conditions with $\nu = 0$. If $|A_{ij}) \le \rho/ \lambda$, then $v^* = 0$ and $t= \rho A_{ij} /\lambda$ satisfies the KKT conditions since $|t| \le 1$. If $|A_{ij}) > \rho/ \lambda$, then the given solution with $t=A_{ij}/|A_{ij}|$ satisfies the KKT conditions. This proves that the solution (\ref{soln}) minimizes $J_{ij}(V_{ij})$.

\begin{algorithm} 
\caption{ADMM Algorithm for Constrained Lasso and Constrained Log-Sum Penalized Log-Likelihood }
\label{alg0}

\algorithmicrequire{\; Number of samples $n$, number of nodes $p$, data $\{{\bm x}(t)\}_{t=1}^{n}$, ${\bm x} \in \mathbb{R}^{p}$, regularization and penalty parameters {$\lambda_0$} and $\rho_0$, tolerances $\tau_{abs}$ and $\tau_{rel}$, variable penalty factor $\mu$, LSP parameter $\epsilon$, maximum number of outer loop iterations $k_{o,max}$, maximum number of inner loop iterations $k_{i,max}$. 

\begin{algorithmic}[1] 
\STATE Calculate sample covariance $\hat{\bm{\Sigma}} = \frac{1}{n} \sum_{t=1}^n {\bm x}(t) {\bm x}^\top(t)$ (after centering ${\bm x}(t)$). 
\STATE Initialize $\lambda_{ij} = \lambda_0$ for every $i \ne j$, $k_o = 1$.
\WHILE{$k_o \le k_{o,max}$,}
\STATE \parbox[t]{\dimexpr\linewidth-1.5em}{Initialize: ${\bm U}^{(0)} = {\bm V}^{(0)} = {\bm 0}$, ${\bm \Omega}^{(0)} = (\mbox{diag}({\hat{\bm \Sigma}}))^{-1}$, where ${\bm U}, {\bm V} \in \mathbb{R}^{p \times p}$, $\rho^{(0)} = \rho_0$. \strut}
\STATE converged = FALSE, $k=0$
\WHILE{converged = FALSE $\;$ AND $\;$ $k \le k_{i,max}$,}
\STATE \parbox[t]{\dimexpr\linewidth-3em}{Eigen-decompose $\hat{\bm \Sigma} - \rho^{(k)} \left( {\bm V}^{(k)} + {\bm U}^{(k)} \right)$ as $\hat{\bm \Sigma} - \rho^{(k)} \left( {\bm V}^{(k)} + {\bm U}^{(k)} \right) = {\bm Q}{\bm D}{\bm Q}^\top$ with diagonal matrix ${\bm D}$ consisting of eigenvalues and orthogonal matrix ${\bm Q}$ consisting of corresponding eigenvectors.  Define diagonal matrix $\tilde{\bm D}$ with $\ell$th diagonal element
$\tilde{\bm D}_{\ell \ell} = ( -{\bm D}_{\ell \ell} + \sqrt{ {\bm D}_{\ell \ell}^2 + 4 \rho^{(k)}  } \, )/(2 \rho^{(k)})$. Set $\bm{\Omega}^{(k+1)} = {\bm Q} \tilde{\bm D} {\bm Q}^\top$. \strut}
\STATE \parbox[t]{\dimexpr\linewidth-3em}{Define thresholding operator $S_{neg}(a, \beta) := (1-\beta/|a|)_+ a_-$ where $(a)_+ := \max(0,a)$ and $(a)_- := \min(0,a)$. The $(i,j)$th element of ${\bm V}$ is updated as in (\ref{soln}):
\begin{align*}
V_{ij}^{(k+1)} & = [ \bm{\Omega}^{(k+1)} - {\bm U}^{(k)} ]_{ii} \quad \mbox{  if } i=j \, , \\ 
V_{ij}^{(k+1)} & = S_{neg} \big( [\bm{\Omega}^{(k+1)} - {\bm U}^{(k)} ]_{ij}, 
             \frac{  \lambda_{ij}}{\rho^{(k)}} \big) 
					     \\
						& \hspace*{0.6in} \mbox{ if } i \neq j \, .
\end{align*} }
\STATE Update ${\bm U}^{(k+1)} = {\bm U}^{(k)} + \left(  {\bm V}^{(k+1)} - \bm{\Omega}^{(k+1)}  \right)$. 
\STATE \parbox[t]{\dimexpr\linewidth-3em}{Check convergence. Set tolerances
\begin{align*}
  & \tau_{pri} =  p \, \tau_{abs} + \tau_{rel} \, \max ( \| {\bm \Omega}^{(k+1)} \|_F , \| {\bm V}^{(k+1)} \|_F ) \\
  & \tau_{dual} =  p \, \tau_{abs} + \tau_{rel} \,  \| {\bm U}^{(k+1)} \|_F / \rho^{(k)} \, .
\end{align*}
Define $d_p = \| {\bm \Omega}^{(k+1)} - {\bm V}^{(k+1)} \|_F$, $d_d = \rho^{(k)} \| {\bm V}^{(k+1)} - {\bm V}^{(k)} \|_F$.
If $( d_p \le \tau_{pri}) \; AND \; (d_d \le \tau_{dual})$, set converged = TRUE \strut}
\STATE \parbox[t]{\dimexpr\linewidth-3em}{Update penalty parameter $\rho$ $\,$ : 
\[
  \rho^{(k+1)} = \left\{ \begin{array}{ll} 2 \rho^{(k)} & \mbox{if  } d_p > \mu d_d \\  
	                                         \rho^{(k)} /2 & \mbox{if  } d_d > \mu d_p \\
																					 \rho^{(k)} & \mbox{otherwise} \, . \end{array} \right.
\]
Set ${\bm U}^{(k+1)} = {\bm U}^{(k+1)}/2$ for $d_p > \mu d_d$ and ${\bm U}^{(k+1)} = 2 {\bm U}^{(k+1)}$ for $d_d > \mu d_p$. \strut}
\STATE $k \leftarrow k+1$
\ENDWHILE
\STATE \parbox[t]{\dimexpr\linewidth-1.5em}{$\hat{\bm \Omega} = {\bm \Omega}^{(k)}$, $k_o \leftarrow k_o+1$, and $\lambda_{ij} = \frac{\lambda}{ | \hat{\Omega}_{ij} | + \epsilon }$ $\forall$ $i \ne j$.  \strut}
\ENDWHILE 
\STATE If $|V_{ij}| > 0$ ($i \ne j$), assign $\{ i,j\} \in \hat{\cal E}$, else $\{ i,j\} \not\in \hat{\cal E}$.
\end{algorithmic}
\algorithmicensure{\;\ $\hat{\bm \Omega} = \bm{\Omega}^{(k)}$, $\hat{\bm W} = -\hat{\bm \Omega}^{-}$, $\hat{\bm L}= \hat{\bm D}-\hat{\bm W}$ and $\hat{\cal E}$} }
\end{algorithm}

A pseudocode for the ADMM algorithm used in this paper is given in Algorithm \ref{alg0} where the outer loop (indexed by $k_o$ in lines 2, 3 and 14 of the code) refers to iterative minimization of $f_{LSP}({\bm \Omega})$ given by (\ref{neq2000a}), and the inner loop (indexed by $k$ in lines 6-12 and 14) refers to minimization of a local linear approximation to $f_{LSP}({\bm \Omega})$, as specified in (\ref{neq2010}).
For (constrained) lasso we take $\lambda_{ij}= \lambda_0$ for all $i \ne j$ in Algorithm \ref{alg0}; this is outer loop iteration $k_o=1$. In subsequent outer loop iterations, we use $\lambda_{ij}$ as specified in line 14 of the code. As implemented in this paper, we run the outer loop for a fixed number of outer iterations, and we obtain excellent results with two outer iterations. One could use a stopping criterion for outer loop also. 

In Algorithm \ref{alg0}, we use the stopping (convergence) criterion following \cite[Sec.\ 3.3.1]{Boyd2010} and varying penalty parameter $\rho$ following \cite[Sec.\ 3.4.1]{Boyd2010}. The stopping criterion is based on primal and dual residuals being small where, in our case, at $(k+1)$st iteration, the primal residual is given by $\bm{\Omega}^{(k+1)} - {\bm V}^{(k+1)}$ and the dual residual by $\rho^{(k)} ({\bm V}^{(k+1)} - {\bm V}^{(k)})$. Convergence criterion is met when the norms of these residuals are below primary and dual tolerances $\tau_{pri}$ and $\tau_{dual}$, respectively; see line 8 of Algorithm \ref{alg0}. In turn, $\tau_{pri}$ and $\tau_{dual}$ are chosen using an absolute and relative criterion as in line 10 of Algorithm \ref{alg0} where $\tau_{abs}$ and $\tau_{rel}$ are user chosen absolute and relative tolerances, respectively. As stated in \cite[Sec.\ 3.4.1]{Boyd2010}, one may use ``possibly different penalty parameters $\rho^{(k)}$
for each iteration, with the goal of improving the convergence in practice, as well as making performance less dependent on the initial choice of the penalty parameter.'' Line 11 of Algorithm \ref{alg0} follows typical choices given in \cite[Sec.\ 3.4.1]{Boyd2010}. 

For all numerical results presented in the paper, we used $\rho_0 =2$, $\mu =10$, $\tau_{abs}=\tau_{rel} =10^{-4}$, and $\epsilon = 10^{-5}$. Furthermore, we used $k_{o,max} =2$, i.e., initialize with constrained lasso and then use one iteration of constrained adaptive lasso.

\section{Theoretical Analysis}  \label{ThA}
In this section we analyze consistency (Theorem 1) and sparsistency (Theorem 2) of the proposed approach under the assumption that ${\bm \Omega} \succ {\bm 0}$; proofs are in the Appendix. For consistency  we follow the method of \cite{Rothman2008} which deals with the lasso penalty.  Dependence of $p$ and $\lambda$ on sample size $n$ is explicitly denoted as $p_n$ and $\lambda_n$, respectively.  

Let $\bm{\Omega}_0 $ denote the true $\bm{\Omega} $ and ${\cal E}_0$ denote the true edgeset $\{ \{i,j\} ~:~ \Omega_{0ij} \ne 0, ~i\ne j \}$. Assume
\begin{itemize}
\setlength{\itemindent}{0.3in}
\item[(A1)] Assume that card$({\cal E}_0) =|{\cal E}_0| \le s_{n0}$.
\item[(A2)] The minimum and maximum eigenvalues of $\bm{\Sigma}_0 = \bm{\Omega}_0^{-1} \succ {\bm 0}$ satisfy 
$0 < \beta_{\min} \le \phi_{\min}(\bm{\Sigma_0}) \le \phi_{\max}(\bm{\Sigma_0}) \le \beta_{\max} < \infty$. 
Here $\beta_{\min}$ and $\beta_{\max}$ are not functions of $n$.
\end{itemize}
Let $\hat{\bm{\Omega}}_\lambda = \arg\min_{\bm{\Omega} \succ {\bm 0}, \; {\bm \Omega} \in {\cal V}_p}  f_{LSP}({\bm \Omega})$.
Theorem 1 establishes local consistency of $\hat{\bm{\Omega}}_\lambda$, i.e., when minimizer is additionally restricted to a ``small'' neighborhood of ${\bm \Omega}_0$. \\
{\it Theorem 1 (Consistency)}: For $\tau > 2$, let
\begin{equation}  \label{naeq58}
   C_0 = 40 \, \max_k ( \Sigma_{0kk}) \sqrt{ 2 \left( \tau + \ln (4) / \ln (p_n) \right) } \, .
\end{equation}
Given real numbers $\delta_1 \in (0,1)$, $\delta_2 > 0$ and $C_1 > 1$, let $C_2 = 1+C_1$, and
\begin{align}  
    M = & (1+\delta_1)^2 (2C_2 + \delta_2) C_0 / \beta_{\min}^2 ,    \label{neq15ab0}  \\
    r_n = & \sqrt{ \frac{(p_n+ s_{n0}) \ln (p_n)}{n}} = o(1)\, , \label{neq15ab1} \\
    N_1 = &  2  \left( \ln (4) + \tau \, \ln ( p_n) \right) ,  \label{neq15ab2} \\
		N_2 = &  \arg \min \left\{ n \, : \, r_n \le \max \Big( \frac{\epsilon (C_1-1)}{M}, \, 
		 \frac{\delta_1}{M \beta_{\min}} \Big)  \right\} \, . \label{neq15ab3}
\end{align}
Suppose the regularization parameter $\lambda_n/\epsilon$ satisfies 
\begin{equation}  
  C_1 C_0 \sqrt{\frac{\ln (p_n)}{n}} \le \frac{\lambda_n}{\epsilon} \le 
     C_1 C_0 \sqrt{ \Big( 1+\frac{p_n}{ s_{n0}} \Big) \frac{\ln ( p_n)}{n} } \, . \label{neq15abc}
\end{equation}
Then if the sample size $n >  \max \{ N_1, N_2 \}$ and assumptions (A1)-(A2) hold true, there exists a local minimizer $\hat{\bm{\Omega}}_\lambda$ such that
\begin{equation}  \label{neq15}
  \| \hat{\bm{\Omega}}_\lambda - \bm{\Omega}_0 \|_F 
	        \le M  r_n
\end{equation}
with prob.\ greater than $1-1/p_n^{\tau-2}$. In terms of rate of convergence,  
  $\| \hat{\bm{\Omega}}_\lambda - \bm{\Omega}_0 \|_F 
	        = {\cal O}_P \left( r_n \right) $ $\quad \bullet$

Sparsistency refers to the property that all parameters that are zero are actually estimated as zero with probability tending to one, as $n \rightarrow \infty$ \cite{Lam2009}. Theorem 2  deals with sparsistency of $\hat{\bm{\Omega}}_\lambda$. Its proof follows that of \cite[Theorem 2]{Lam2009} pertaining to lasso and SCAD penalties

{\it Theorem 2 (Sparsistency)}: Suppose Theorem 1 holds true so that (\ref{neq15}) holds. In addition, suppose that there exists a sequence $\eta_n \rightarrow 0$ such that $\| \hat{\bm{\Omega}}_\lambda - \bm{\Omega}_0 \| = {\cal O}_P ( \eta_n)$ and $\sqrt{\ln(p_n)/n} + \eta_n = {\cal O}(\lambda_n)$. Then with prob.\ tending to one, $\hat{{\Omega}}_{\lambda ij} = 0$ for all $\{i,j\} \in {\cal E}_0^c =\{ \{i,j\} ~:~ \Omega_{0ij} = 0, ~i\ne j \}$. $\quad \bullet$

{\it Remark 1}: For both consistency and sparsistency to be satisfied, the chosen regularization parameters $\lambda_n$'s need to be compatible. Theorem 1 imposes  upper and lower bounds on the rate of $\lambda_n$ and Theorem 2 specifies a lower bound. Therefore, for both  consistency and sparsistency to be satisfied, we must have
\begin{equation}  \label{neq15com}
     \sqrt{\ln(p_n)/n} + \eta_n \asymp \lambda_n/\epsilon \asymp  \sqrt{ \Big( 1+\frac{p_n}{ s_{n0}} \Big) 
		    \frac{\ln ( p_n)}{n} } .
\end{equation}
Its consequences depend upon $\eta_n$ required to attain $\| \hat{\bm{\Omega}}_\lambda - \bm{\Omega}_0 \| = {\cal O}_P ( \eta_n)$. As discussed in \cite{Lam2009} for lasso, we consider two cases, using the inequalities $\|{\bm A}\|_F/\sqrt{p_n} \le \|{\bm A}\| \le \|{\bm A}\|_F$ for ${\bm A} \in \mathbb{R}^{p_n \times p_n}$. 
\begin{itemize}
\item[(a)] Since $\| \hat{\bm{\Omega}}_\lambda - \bm{\Omega}_0 \| \le \| \hat{\bm{\Omega}}_\lambda - \bm{\Omega}_0 \|_F$, in the worst case where the two have the same order, $\| \hat{\bm{\Omega}}_\lambda - \bm{\Omega}_0 \|$ =  ${\cal O}_P \left( \sqrt{ \frac{(p_n+s_{n0}) \ln (p_n) }{n} } \right)$ so that $\eta_n = \sqrt{ \frac{ (p_n+s_{n0}) \ln (p_n) }{n} }$. Then for (\ref{neq15com}) to hold true, we should have $1+  \sqrt{p_n+s_{n0}} \asymp  \sqrt{ 1+(p_n/(s_{n0}))}$, which 
holds only if $s_{n0} = {\cal O}(1)$.

\item[(b)] Since $\| \hat{\bm{\Omega}}_\lambda - \bm{\Omega}_0 \|_F/\sqrt{p_n} \le \| \hat{\bm{\Omega}}_\lambda - \bm{\Omega}_0 \|$, in the optimistic case where the two have the same order, $\| \hat{\bm{\Omega}}_\lambda - \bm{\Omega}_0 \| = {\cal O}_P \left( \sqrt{ (1+\frac{s_{n0}}{p_n}) \ln \frac{p_n}{n}} \right)$ so that $\eta_n = \sqrt{ (1+\frac{s_{n0}}{p_n}) \ln \frac{p_n}{n}}$. Then for (\ref{neq15com}) to hold true, we should have $1+\sqrt{1+\frac{s_{n0}}{p_n}} \asymp  \sqrt{ 1+\frac{p_n}{s_{n0}}}$, which holds only if $s_{n0} = {\cal O}(p_n)$.
\end{itemize}

{\it Remark 2}: Results of \cite{Loh2015, Loh2017} are related. In \cite[Sec.\ 3.4]{Loh2015} Theorem 1 type results for any stationary point of graphical lasso under $\ell_1$, SCAD and MCP penalties are shown, however, \cite{Loh2015} does not discuss LSP. Our results are for a ``local'' stationary point but for LSP. Moreover \cite[Sec.\ 3.4]{Loh2015} considers graphical lasso where  penalties are applied to all terms of ${\bm \Omega}$ whereas we apply LSP only to off-diagonal ${\bm \Omega}$. LSP (along with other non-convex penalties such as SCAD and MCP) applied only to off-diagonal terms of ${\bm \Omega}$ is considered in \cite[Appendix E]{Loh2017} for Theorem 2 type results (support recovery). \cite[Appendix E]{Loh2017} shows that an incoherence condition is required for support recovery for $\ell_1$ and LSP penalties, but not for SCAD and MCP. In our Theorem 2 we do not need any such incoherence condition. 

\begin{figure*}[t]
\centerline{\includegraphics[width=1.0\textwidth]{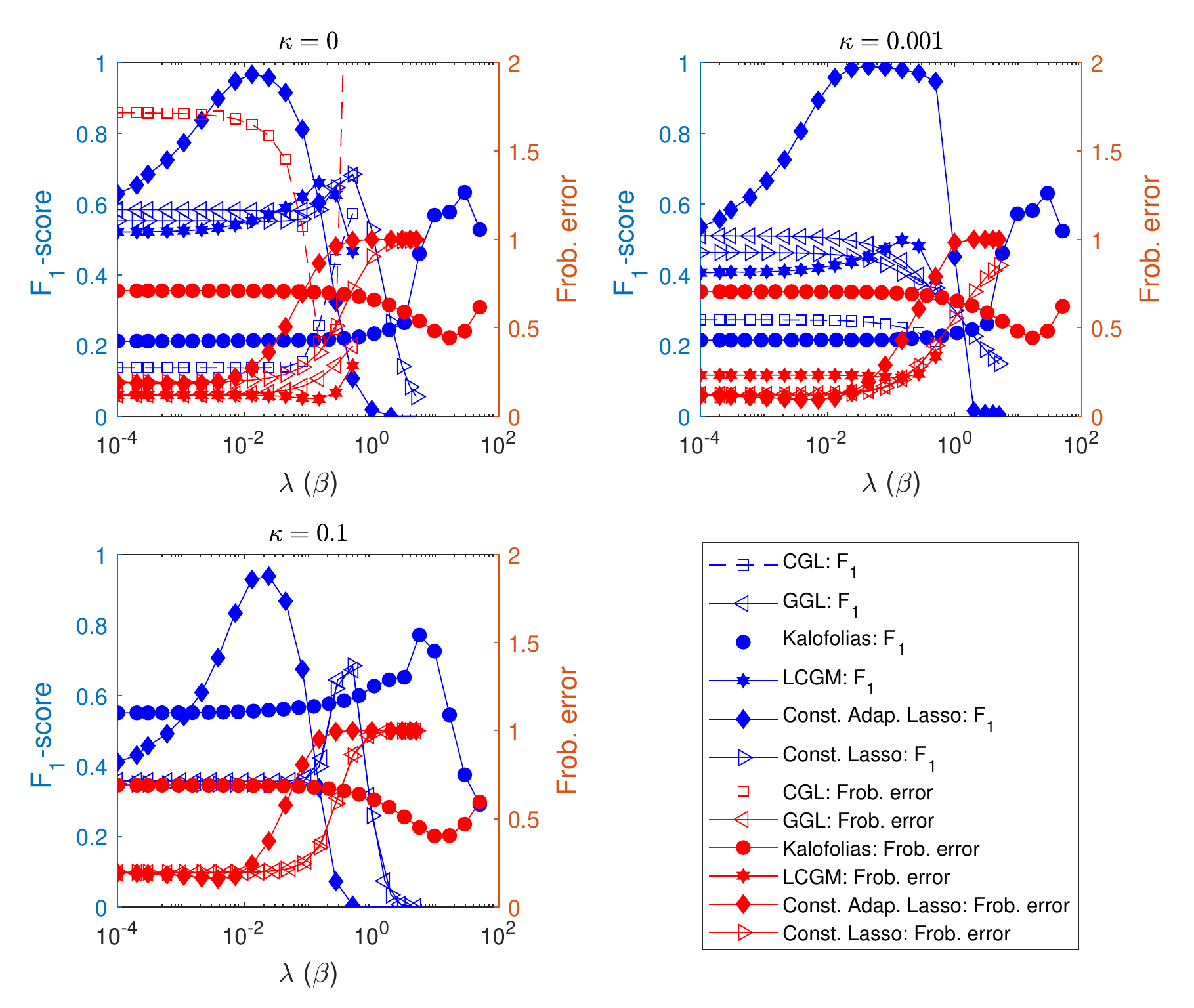}} 
\caption{Erd\"{o}s-R\`{e}nyi graph ${\cal G}_{ER}^{100,0.03}$, $p=100$, $n=400$, 100 runs: $F_1$ score and weight matrix error norm. CGL and GGL are from \cite{Egilmez2017}, Kalofolias is \cite{Kalofolias2016}, LCGM is \cite{Ying2020}, Const.\ Lasso and Const.\ Adap.\ Lasso are proposed approaches, using ADMM. } \label{fig1}
\end{figure*}

\section{Numerical Results} \label{SE}
We now present numerical results for both synthetic and real data to illustrate the proposed approach. In synthetic data examples the ground truth is known and this allows for assessment of the efficacy of various approaches. In real data examples where the ground truth is unknown, our goal is visualization and exploration of the dependency structures underlying the data, similar to \cite{Danaher2014, Pavez2016, Pavez2018, Ying2020}.

\subsection{Synthetic Data}
 We consider two Gaussian graphical models: a chain graph where $p$ nodes are connected in succession, and an Erd\"{o}s-R\`{e}nyi graph ${\cal G}_{ER}^{p,p_{er}}$ where $p$ nodes are connected with probability $p_{er} =0.03$. In each model, in the upper triangular $\Omega$ (inverse covariance), $\Omega_{ij} =0$ if $\{i,j\} \not\in {\cal E}$, and $\Omega_{ij}$ is uniformly distributed over $[-0.3,-0.1]$ if $\{i,j\} \in {\cal E}$. With $\Omega = \Omega^\top$, we take $\Omega_{ii}=-\sum_{j=1}^p \Omega_{ij}$ for every $i$, yielding the combinatorial Laplacian ${\bm L} = {\bm \Omega}$. Now add $\kappa {\bm I}$ to  ${\bm{\Omega}}$ with $\kappa$ picked to make minimum eigenvalue of ${\bm{\Omega}} + \kappa {\bm I}$ equal to 0, 0.001 or 0.1, and with $\bm{\Phi} \bm{\Phi}^\top = \left( {\bm{\Omega}} + \kappa {\bm I} \right)^\dagger$, we generate ${\bm x} = \bm{\Phi} {\bm w}$ with ${\bm w} \in \mathbb{R}^{p}$ as Gaussian ${\bm w} \sim {\mathcal N}( {\bm 0}, \bm{I})$. We generate $n$ i.i.d.\ observations from ${\bm x}$ using $p =100$. Addition of $\kappa {\bm I}$, $\kappa > 0$, yields a generalized Laplacian matrix ${\bm L}_g = {\bm L} + \kappa {\bm I}$ \cite{Egilmez2017}. 

While single-component combinatorial Laplacian matrix for the chain graph is always connected (i.e., degree of each node is at least one), that for a $p$-node  Erd\"{o}s-R\`{e}nyi graph ${\cal G}_{ER}^{p,p_{er}}$ may not always be so, particularly if the probability $p_{er}$ of any two nodes being connected is low. In our simulation $p_{er} = 0.03$. Therefore, in each run, we checked if every node had a degree $\ge 1$. If not, we randomly connected an unconnected node to one of the other $p-1$ nodes (with $\Omega_{ij}$ uniformly distributed over $[-0.3,-0.1]$).

\begin{figure*}[t]
\centerline{\includegraphics[width=1.0\textwidth]{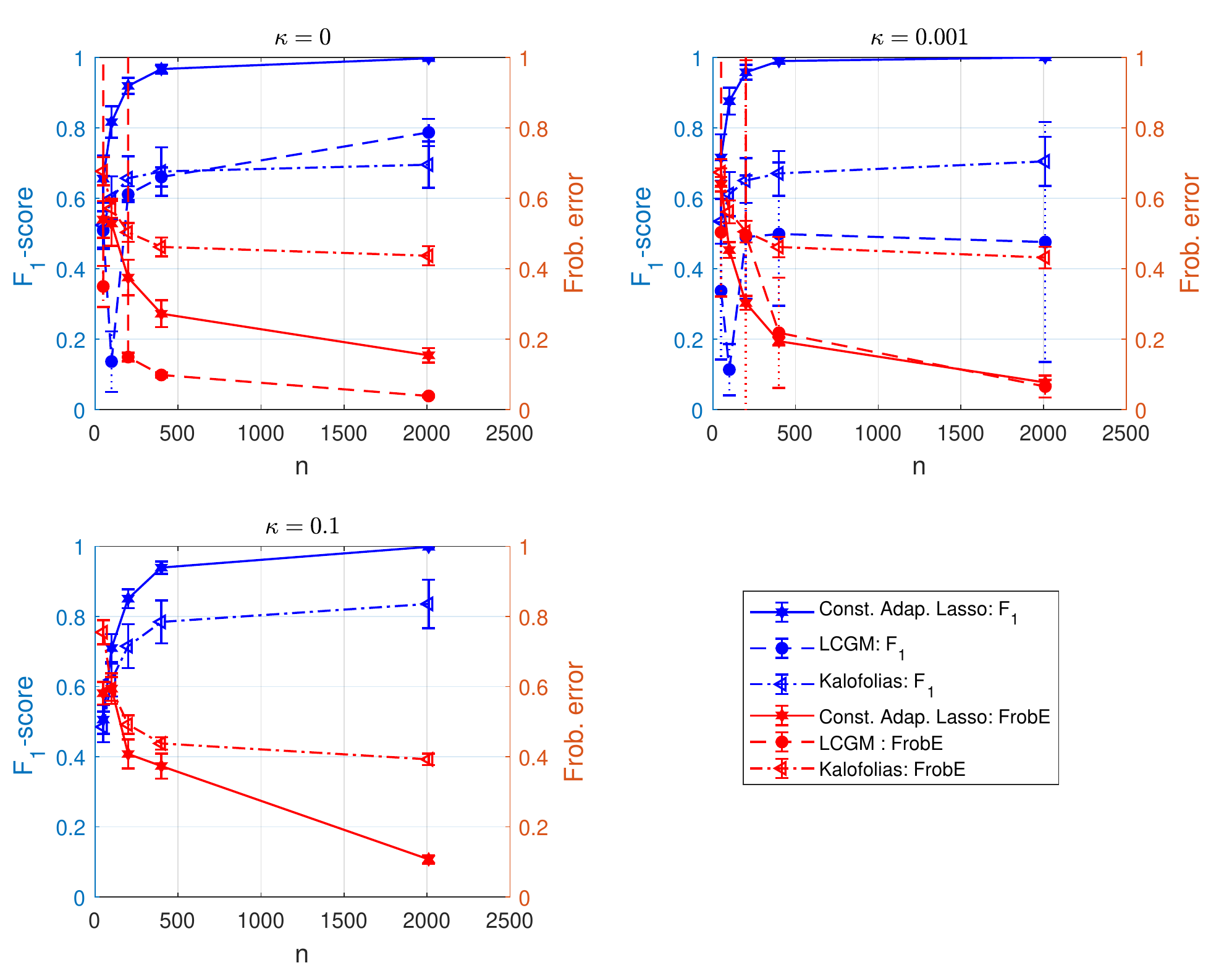}}
\caption{Erd\"{o}s-R\`{e}nyi graph ${\cal G}_{ER}^{100,0.03}$, $p=100$, 100 runs: $F_1$ score and weight matrix error norm. Kaloflias is \cite{Kalofolias2016}, LCGM is \cite{Ying2020}, Const.\ Adap.\ Lasso is proposed adaptive lasso. } \label{fig2}
\end{figure*}

We apply six methods for estimating the true edgeset ${\cal E}_0$ and true off-diagonal ${\bm \Omega}_0^-$: 
\begin{itemize} 
\item[(i)] Combinatorial graph Laplacian (CGL) method of \cite{Egilmez2017}, using MATLAB function estimate\_cgl.m from \cite{Egilmez2017b}.
\item[(ii)] Generalized graph Laplacian (GGL) method of \cite{Egilmez2017}, using MATLAB function estimate\_ggl.m from \cite{Egilmez2017b}
\item[(iii)] The signal smoothness-based method of \cite{Kalofolias2016} which yields the weighted adjacency matrix ${\bm W}$, equaling $-{\bm \Omega}_0^-$. The matlab software is available in \cite{Perraudin2016}.
\item[(iv)]  Laplacian-constrained graphical model (LCGM) method of \cite{Ying2020} with MCP penalty, using R-implementation cited in \cite{Ying2020}
\item[(v)] Our proposed constrained adaptive lasso (CAL) method with LSP ($\lambda_{ij} = \lambda/(|\bar{\Omega}_{ij}| + \epsilon)$, $\epsilon=10^{-5}$).
\item[(vi)] Our proposed constrained lasso (CL) method ($\lambda_{ij} = \lambda$), the solution of which taken to be  $\bar{\Omega}$ for the CAL approach.
\end{itemize}
The performance measures are $F_1$-score $\in [0,1]$ for efficacy in edge detection (higher is better), and normalized Frobenius error norm in estimating ${\bm \Omega}_0^-$ (off-diagonal true ${\bm \Omega}_0$) (lower is better), defined as
$\| c \hat{\bm \Omega}^- - {\bm \Omega}_0^-\|_F  / \| {\bm \Omega}_0^-\|_F $ where $c=1$ for all approaches except \cite{Kalofolias2016} which yields ${\bm W}$ up to a scale factor, therefor, $c$ is picked to minimize the error norm in that case. The $F_1$-score is defined as 
\begin{align*}
F_1 =  & \frac{ 2 \times \mbox{precision} \times \mbox{recall} } { \mbox{precision} + \mbox{recall} } \, , \\
\mbox{precision} = & \frac{  | \hat{\cal E} \cap {\cal E}_0| }{ |\hat{\cal E}| } \, , \quad
 \mbox{recall} = \frac{  |\hat{\cal E} \cap {\cal E}_0| }{ |{\cal E}_0| }
\end{align*}
and ${\cal E}_0$ and $ \hat{\cal E}$ denote the true and estimated edge sets, respectively.

In Fig.\ \ref{fig1} for a sample size $n=400$, we show the performance as a function of penalty parameter $\lambda$ ($\lambda/\epsilon$ for adaptive lasso, and $\beta$ for \cite{Kalofolias2016}) for $\kappa \in \{0, 0.001, 0.1\}$: higher $\lambda$ (lower $\beta$ for  \cite{Kalofolias2016}) should lead to sparser graphs. CGL and LCGM are designed specifically for $\kappa =0$ whereas other approaches do not crucially depend on it; we did not implement CGL and LCGM for $\kappa=0.1$.  As noted in \cite{Ying2020, Zhang2020}, Laplacian-constrained log-likelihood approaches do not yield sparse graphs under convex penalties; we see this in the performance of CGL: it was not implemented for $\kappa = 0.1$, and for $\kappa = 0.001$ its Frobenius error norm is off the graph ($> 17$). Our proposed CAL has the best $F_1$ performance for all values of $\kappa$: for a choice of some $\lambda$, $F_1$ score exceeds 0.95 whereas other approaches do not perform nearly as well. \cite{Kalofolias2016} performs better than other approaches except proposed CAL, in terms of $F_1$ score, but has poor Frobenius error performance (as was noted in \cite{Egilmez2017}). In Fig.\ \ref{fig2} we show performance for sample sizes $n=50,100,200,400,2000$ for \cite{Kalofolias2016}, \cite{Ying2020} and proposed constrained adaptive lasso (CAL), where penalty parameters were optimized for best $F_1$ scores. LCGM with MCP penalty \cite{Ying2020} (not implemented for $\kappa = 0.1$) has some numerical conditioning problem for $n=100$, resulting in poor $F_1$ score and excessive Frobenius error (such phenomenon has also been noted in \cite{Zhang2020}). We see that the proposed CAL performs the best independent of $\kappa$ value, while LCGM sharply deteriorates for $\kappa > 0$.

\begin{figure}[hbt]
\centerline{\includegraphics[width=0.5\textwidth]{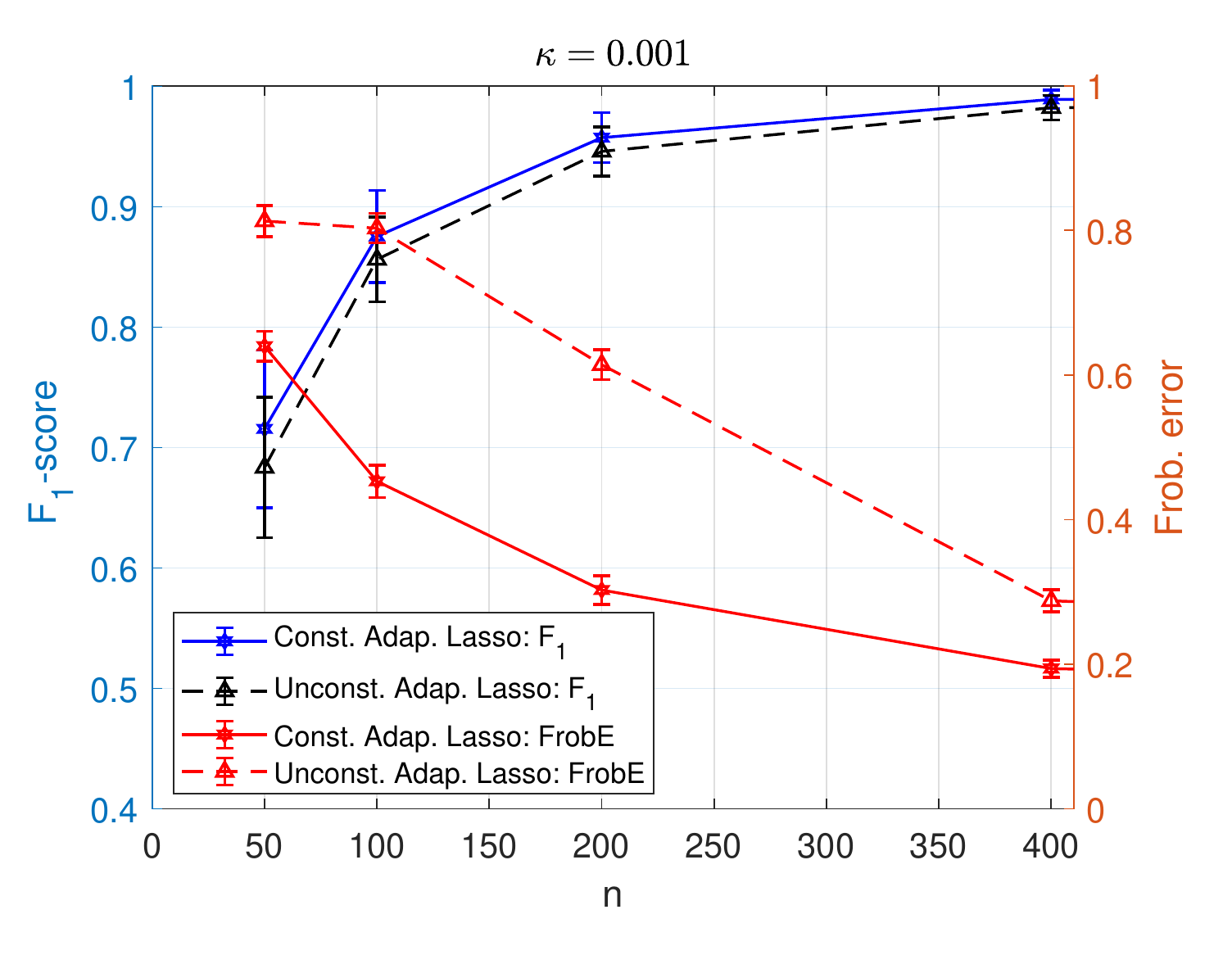}} 
\caption{Erd\"{o}s-R\`{e}nyi graph ${\cal G}_{ER}^{100,0.03}$, $p=100$, 100 runs: $F_1$ score and weight matrix error norm for  Const.\ Adap.\ Lasso (proposed) and unconstrained Adap.\ Lasso approaches, using ADMM. The unconstrained Adap.\ Lasso approach does not impose constraint (\ref{neq130}).} \label{fig3}
\end{figure}

In Fig.\ \ref{fig3} we present a comparison between the results of optimization of $f_{LSP}({\bm \Omega})$ with and without the constraint (\ref{neq130}), labeled ``Const.\ Adap.\ Lasso'' and ``Unconst.\ Adap.\ Lasso,'' respectively. The results are for Erd\"{o}s-R\`{e}nyi graph ${\cal G}_{ER}^{100,0.03}$, $p=100$, $n \in \{50, 100, 200, 400 \}$, based on 100 runs. The largest performance improvement due to imposition of the constraint (\ref{neq130}) is in estimation of ${\bm \Omega}$ (equivalently ${\bm W}$), which is not surprising since ${\bm \Omega}_0 \in {\cal V}_p$.

In Table \ref{table1} we present results for a fixed sample size $n=400$ with varying graph size $p \in \{100, 200, 400, 1000, 2000\}$ to illustrate performance with scaling of the problem size. The results are for Erd\"{o}s-R\`{e}nyi graph ${\cal G}_{ER}^{p,3/p}$ where two nodes are connected with probability $p_{er} = 3/p$ (as in \cite{Kalofolias2016}), and the for data generation we used $\kappa =0$. We show the $F_1$-score and average time per run for four approaches, proposed CAL, GGL \cite{Egilmez2017}, LCGM \cite{Ying2020} and signal smoothness-based method \cite{Kalofolias2016}, where penalty parameters were optimized for best $F_1$ scores. All algorithms were run on a Window Home 10 operating system with processor Intel(R) Core(TM) i5-6400T CPU @2.20 GHz with 12 GB RAM, and all MATLAB implementations were run on MATLAB R2020b. The shown results are based on 10 runs only as time per run increases significantly for larger values of $p$. It is seen that while the proposed CAL method is most demanding computationally, its $F_1$-score performance is the best by a wide margin.

In Table \ref{table2} we show results for the chain graph, corresponding to Fig.\ \ref{fig2}, for sample sizes $n=50,100,200,400$ and $\kappa \in \{0, 0.001\}$. The discussion pertaining to Fig.\ \ref{fig2} applies here as well.

\begin{figure*}[t]
\begin{subfigure}{.33\textwidth}
  \centering
  \includegraphics[width=1.0\linewidth]{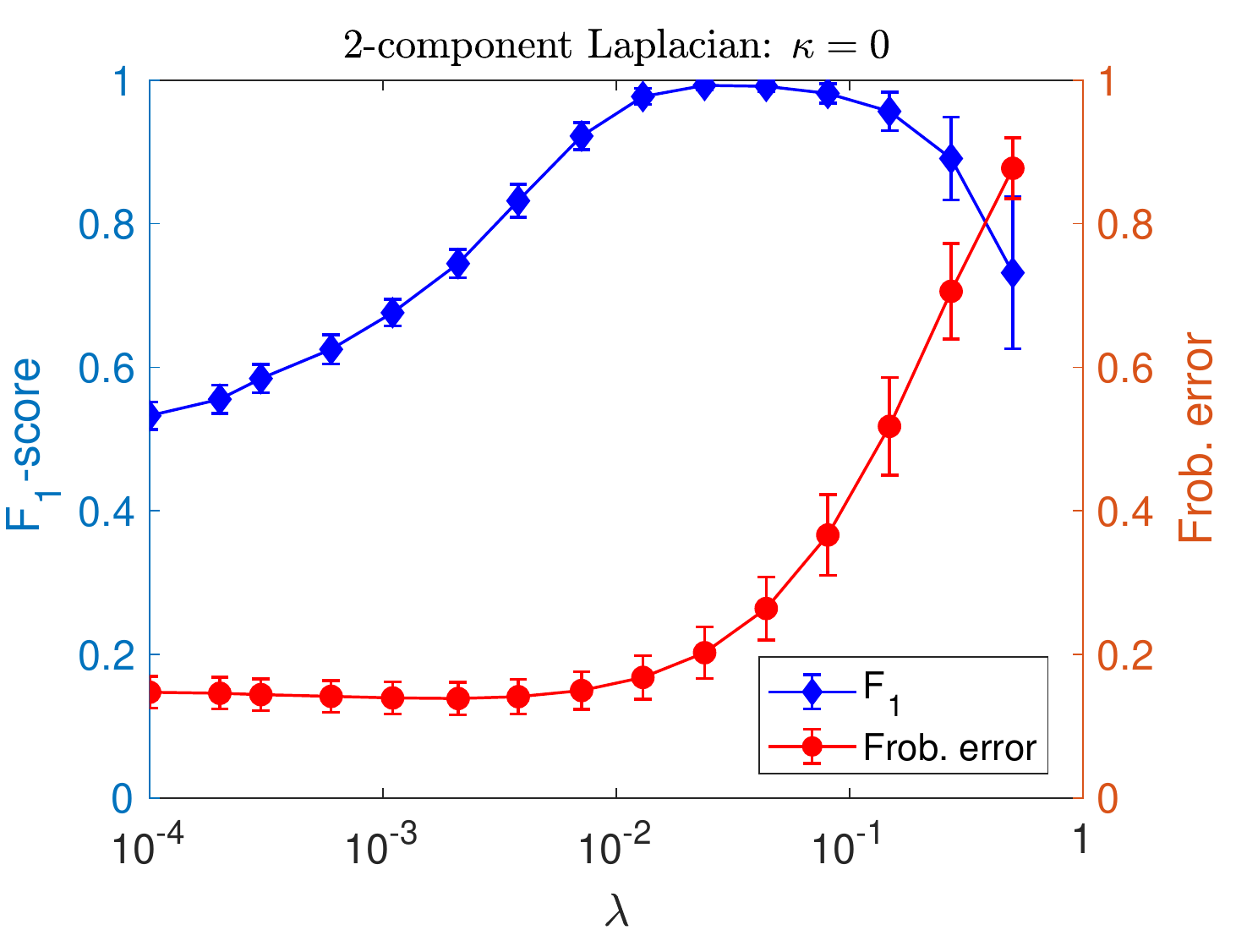}  
  \caption{$F_1$-score and Frobenius error}
  \label{fig4a}
\end{subfigure}%
\begin{subfigure}{.33\textwidth}
  \centering
  \includegraphics[width=1.0\linewidth]{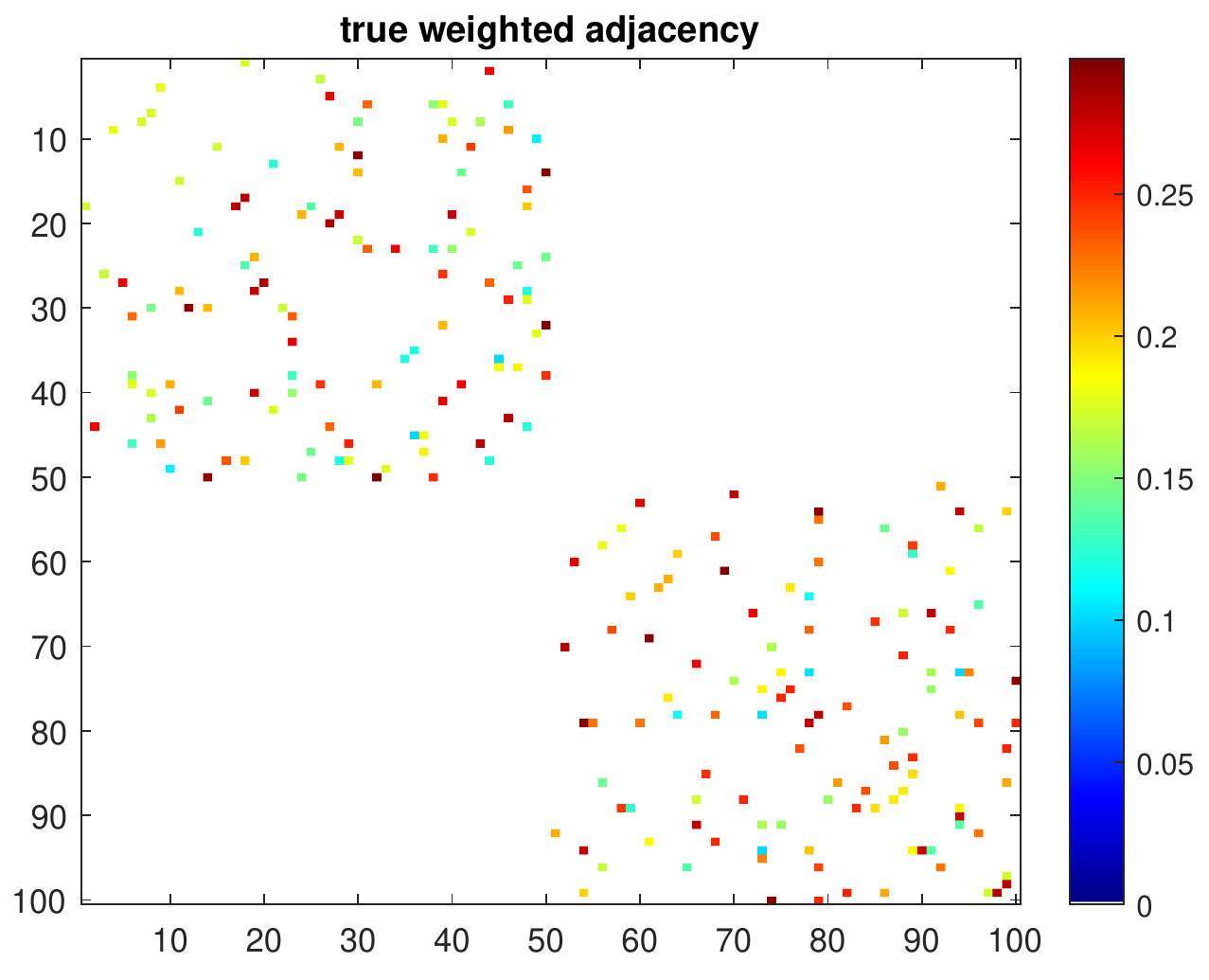}  
  \caption{True $W_{ij}=-\Omega_{ij}$ .}
  \label{fig4b}
\end{subfigure}%
\begin{subfigure}{.33\textwidth}
  \centering
  \includegraphics[width=1.0\linewidth]{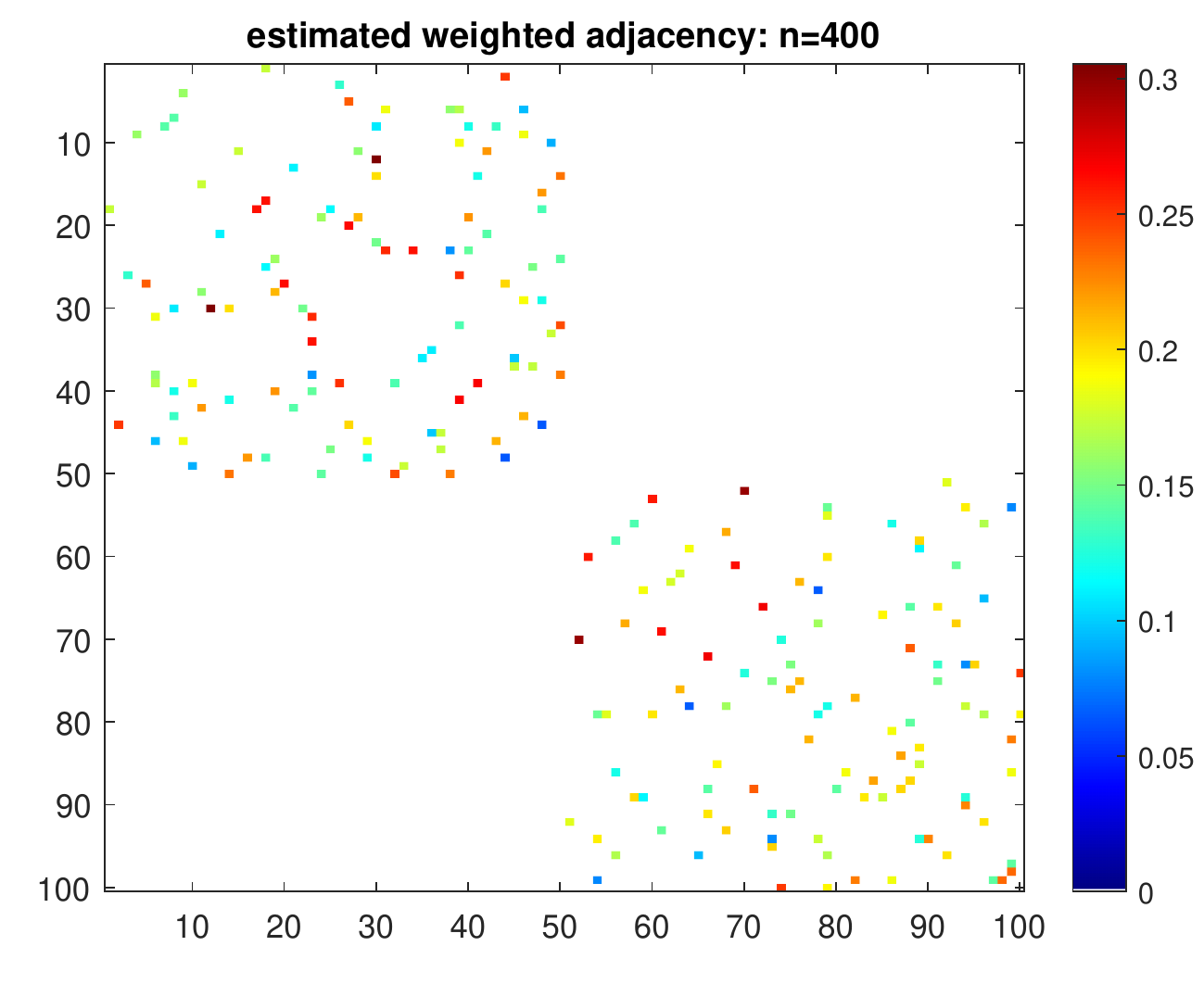}  
  \caption{Estimated $W_{ij}=-\Omega_{ij}$, n=400}
  \label{fig4c}
\end{subfigure}
\caption{Two-component Laplacian, $\kappa=0$, $p=100$, each component is Erd\"{o}s-R\`{e}nyi ${\cal G}_{ER}^{50,0.03}$. (a) performance measures based on 100 runs, (b) sample ${\bm W}$, (c) estimate $\hat{\bm W}$ using proposed CAL. }
\label{fig4}
\end{figure*}

In Fig.\ \ref{fig4a}, for $n=400$ and $p=100$, we show the performance of our proposed CAL approach as a function of $\lambda/\epsilon$, when applied to a two-component Laplacian precision matrix (two zero eigenvalues), $\kappa=0$, $p=100$, each component is independent Erd\"{o}s-R\`{e}nyi ${\cal G}_{ER}^{50,0.03}$. We see that our approach works well (whereas LCGM \cite{Ying2020} is designed only for single-component Laplacians: one zero eigenvalue). In Figs.\ \ref{fig4b} and \ref{fig4c} we show the true and estimated weighted adjacency matrices for a single run using the $\lambda$ value from Fig.\ \ref{fig4a} that maximizes the $F_1$ score.

\begin{table*}
\begin{center}
\begin{tabular}{cccccc}   \hline\hline
 {\bf Model}: & \multicolumn{5}{c}{ {\bf Erd\"{o}s-R\`{e}nyi Graph}: sample size $n$=400 } \\ \hline\hline
 \mbox{number of nodes $p$}  & 100 & 200 & 400 & 1000 & 2000 \\  \hline
  Approach & \multicolumn{5}{c}{ $\kappa = 0$: $F_1$ score ($\pm \sigma$)}  \\  \hline
Kalofolias \cite{Kalofolias2016} & 0.6850 $\pm$0.0578& 0.7991 $\pm$0.0276& 0.7938 $\pm$0.0226& 0.7531 $\pm$0.0107 & *** \\
GGL \cite{Egilmez2017} & 0.6720 $\pm$0.0610& 0.6589 $\pm$0.0446& 0.6168 $\pm$0.0242& 0.5528 $\pm$0.0209 & 0.4932 $\pm$0.0168 \\
LCGM \cite{Ying2020} & 0.6540 $\pm$0.0202& 0.5906 $\pm$0.0152& 0.0438 $\pm$0.0138& *** & *** \\
Const.\ Adap.\ Lasso & 0.9666 $\pm$0.0128& 0.9741 $\pm$0.0053& 0.9651 $\pm$0.0103 & 0.9615 $\pm$0.0057 & 0.9552 $\pm$0.0049  \\ \hline
  & \multicolumn{4}{c}{ $\kappa = 0$: Time (s) ($\pm \sigma$)}  \\  \hline
Kalofolias \cite{Kalofolias2016} & 0.2602 $\pm$0.0028& 0.7182 $\pm$0.0163& 2.5862 $\pm$0.0126& 29.160 $\pm$0.1752 & *** \\
GGL \cite{Egilmez2017} & 0.0300 $\pm$0.0014& 0.1050 $\pm$0.0038& 1.2853 $\pm$0.0493& 19.2271 $\pm$0.15479 & 153.01 $\pm$7.2441 \\
LCGM \cite{Ying2020} & 4.5135 $\pm$0.0591& 31.490 $\pm$7.1055& 9.379 $\pm$1.778& *** & *** \\
Const.\ Adap.\ Lasso & 0.6603 $\pm$0.1212& 4.716 $\pm$0.4591& 17.80 $\pm$1.943 & 130.28 $\pm$6.890 & 866.87 $\pm$20.80\\ \hline
\hline 
\end{tabular} 
\end{center}
\caption{\it $F_1$-score and timing results for Erd\"{o}s-R\`{e}nyi graph ${\cal G}_{ER}^{p,3/p}$ based on 10 runs. Sample size $n$=400, variable graph size $p$. Entry *** indicates that the algorithm failed to converge in 1000 sec.} \label{table1}
\end{table*}

\subsubsection{Model Selection}
In practice, one would select  $\lambda$ via cross-validation or an information criterion.
For selection of $\lambda$, we use the Bayesian information criterion (BIC) 
\[
 \mbox{BIC}(\lambda ) = \mbox{tr} ( \hat{\bm{\Sigma}} \hat{\bm{\Omega}}  ) - \ln(|\hat{\bm{\Omega}}|) 
		     +  \frac{\ln (n)}{n} \mbox{card}( \hat{{\cal E}} )
\]
based on optimized  $- \ln  f_{{\bm X}}({\bm X}) \propto  \frac{n}{2} \big( \mbox{tr} ( \hat{\bm{\Sigma}}  \hat{\bm{\Omega}}  ) - \ln | \hat{\bm{\Omega}}| \big)$.  The tuning parameter $\lambda$ is selected over a grid of values to minimize BIC. 
We search over $\lambda$ values in the range $[\lambda_\ell , \lambda_u]$ selected via the following heuristic. We first find the smallest $\lambda$, labeled $\lambda_{sm}$, for which we get a no-edge model (i.e., $| \hat{\cal E} | =0$). Then we set $\lambda_u = \lambda_{sm}$ and $\lambda_\ell = \lambda_u/100$ for synthetic data, search over 10 logarithmically spaced values in $[\lambda_\ell , \lambda_u]$. The given choice of $\lambda_u$ precludes ``extremely'' sparse models while that of $\lambda_\ell$ precludes ``very'' dense models. The results based on 20 Monte Carlo runs are shown in Table \ref{table3} for Chain and Erd\"{o}s-R\`{e}nyi graphs with $p=100$ and $\kappa =0$, for two sample sizes: $n$=200 and 2000. The BIC-based selected $\lambda$ was used in each run to estimate ${\bm \Omega}$ and compute $F_1$-score and Frobenius error norm. The proposed approach seems to work well. No such approaches are available in \cite{Kalofolias2016, Egilmez2017, Ying2020, Zhang2020}.

\begin{table*}
\begin{center}
\begin{tabular}{ccccc}   \hline\hline
 {\bf Model}: & \multicolumn{4}{c}{ {\bf Chain Graph}: number of nodes $p$=100}  \\ \hline\hline
 \mbox{sample size $n$}  & 50 & 100 & 200 & 400 \\  \hline
  Approach & \multicolumn{4}{c}{ $\kappa = 0$: $F_1$ score ($\pm \sigma$)}  \\  \hline
Kalofolias \cite{Kalofolias2016} & 0.6526 $\pm$0.0425& 0.6546 $\pm$0.0383& 0.6591 $\pm$0.0424& 0.6547 $\pm$0.0335\\
LCGM \cite{Ying2020} & 0.7086 $\pm$0.0367& 0.2410 $\pm$0.1548& 0.6717 $\pm$0.1989& 0.6406 $\pm$0.2086\\
Const.\ Adap.\ Lasso & 0.9905 $\pm$0.0072& 0.9990 $\pm$0.0025& 1.0000 $\pm$0.000& 1.0000 $\pm$0.000 \\ \hline
  & \multicolumn{4}{c}{ $\kappa = 0$: Frobenius Error Norm ($\pm \sigma$)}  \\  \hline
Kalofolias \cite{Kalofolias2016} & 0.6717 $\pm$0.0314& 0.6470 $\pm$0.0303& 0.6220 $\pm$0.0318& 0.6101 $\pm$0.0279\\
LCGM \cite{Ying2020} & 0.1923 $\pm$0.0270& 145.32 $\pm$144.53& 0.2526 $\pm$0.1731& 0.1301 $\pm$0.0491\\
Const.\ Adap Lasso & 0.4410 $\pm$0.0133& 0.4464 $\pm$0.0098& 0.2916 $\pm$0.0078& 0.1165 $\pm$0.0065\\ \hline
Approach & \multicolumn{4}{c}{ $\kappa = 0.001$: $F_1$ score ($\pm \sigma$)}  \\  \hline
Kalofolias \cite{Kalofolias2016} & 0.6766 $\pm$0.0425& 0.6546 $\pm$0.0383& 0.6591 $\pm$0.0424& 0.6789 $\pm$0.0325\\
LCGM \cite{Ying2020} & 0.6478 $\pm$0.0187& 0.1539 $\pm$0.0773& 0.5723  $\pm$0.1240& 0.5609 $\pm$0.0550\\
Const.\ Adap.\ Lasso & 0.9887 $\pm$0.0076 & 0.9988 $\pm$0.0028 & 0.9999 $\pm$0.0007& 1.0000 $\pm$0.000 \\ \hline
  & \multicolumn{4}{c}{ $\kappa = 0.001$: Frobenius Error Norm ($\pm \sigma$)}  \\  \hline
Kalofolias \cite{Kalofolias2016} & 0.6526 $\pm$0.0318& 0.6244 $\pm$0.0299& 0.5998 $\pm$0.0319& 0.5881 $\pm$0.0274\\
LCGM \cite{Ying2020} & 0.1958  $\pm$0.0250& 554.62 $\pm$598.50& 0.1487 $\pm$0.1065& 0.1077 $\pm$0.0308\\
Const.\ Adap Lasso & 0.4415 $\pm$0.0134& 0.4458 $\pm$0.0098& 0.2891 $\pm$0.0078& 0.1778 $\pm$0.0060\\ \hline
\hline
\end{tabular} 
\end{center}
\caption{\it Results for Chain graph based on 100 runs.} \label{table2}
\end{table*}

\begin{table*}
\begin{center}
\begin{tabular}{ccccc}   \hline\hline
 \mbox{sample size $n$}  & 50 & 100 & 200 & 2000 \\  \hline
  Model & \multicolumn{4}{c}{ $\kappa = 0$: $F_1$ score ($\pm \sigma$)}  \\  \hline
Chain Graph & 0.8949 $\pm$0.0268&  0.9581 $\pm$0.0122 & 0.9908 $\pm$0.0077&  1.0000 $\pm$0.000 \\ 
Erd\"{o}s-R\`{e}nyi Graph & 0.6379 $\pm$0.0877&  0.7994 $\pm$0.0606 & 0.9116 $\pm$0.0182&  0.9941 $\pm$0.0121 \\  \hline
  & \multicolumn{4}{c}{ $\kappa = 0$: Frobenius Error Norm ($\pm \sigma$)}  \\  \hline
Chain Graph & 0.1922 $\pm$0.0169&  0.1343 $\pm$0.0094 & 0.1050 $\pm$0.0085&  0.0834 $\pm$0.0026 \\
Erd\"{o}s-R\`{e}nyi Graph & 0.6988 $\pm$0.0789&  0.4959 $\pm$0.1053 & 0.3547 $\pm$0.0664&  0.1981 $\pm$0.0653\\ \hline
\hline
\end{tabular} 
\end{center}
\caption{\it Tuning parameter selection  for proposed constrained adaptive lasso: 20 runs} \label{table3}
\end{table*}

\subsection{Real data: Financial Time Series} 
\begin{figure*}[ht]
\begin{subfigure}[t]{.5\textwidth}
  \centering
  \includegraphics[width=.8\linewidth]{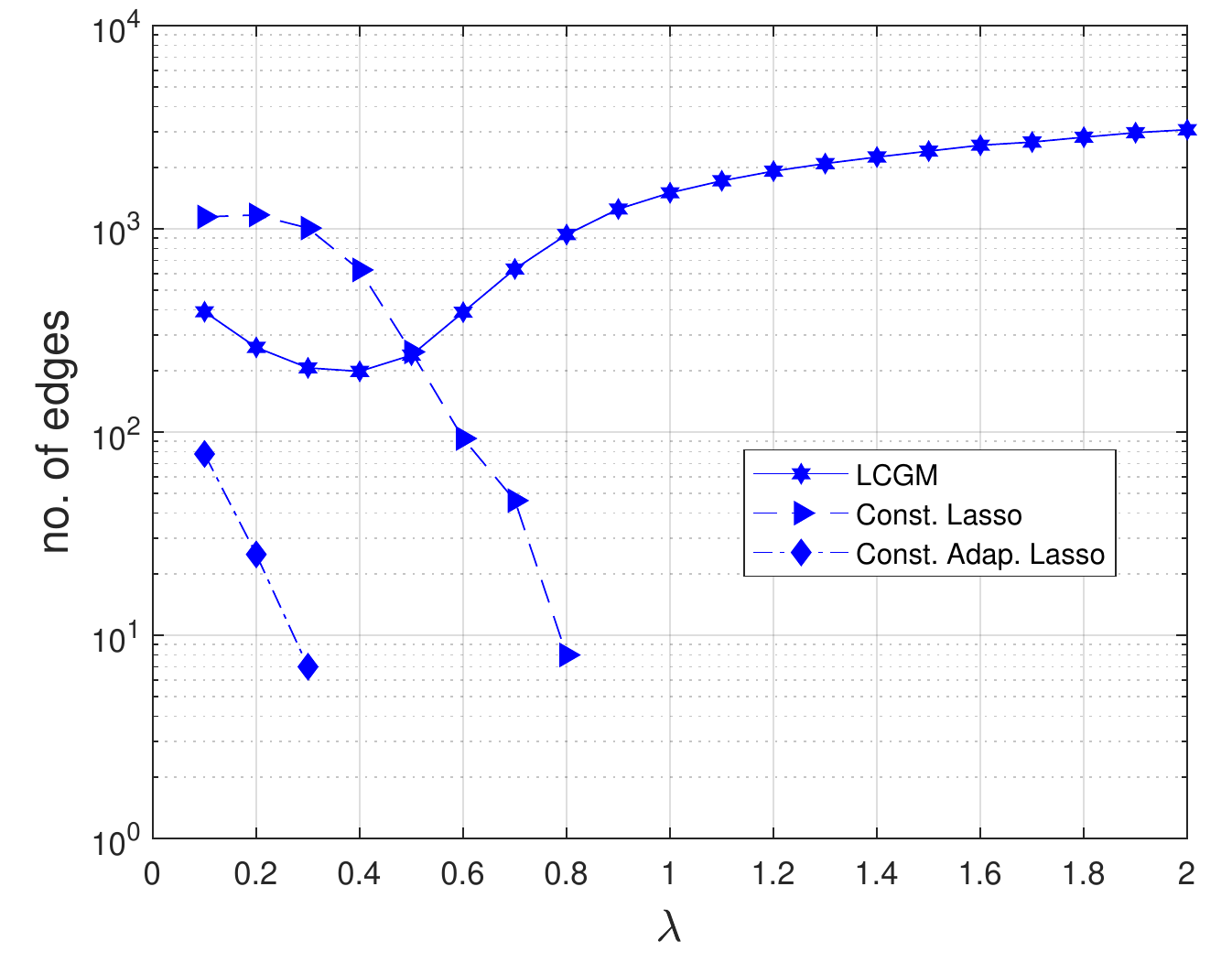}  
  \caption{Number of edges as a function of $\lambda$}
  \label{fig5a}
\end{subfigure}%
\begin{subfigure}[t]{.5\textwidth}
  \centering
  \includegraphics[width=.8\linewidth]{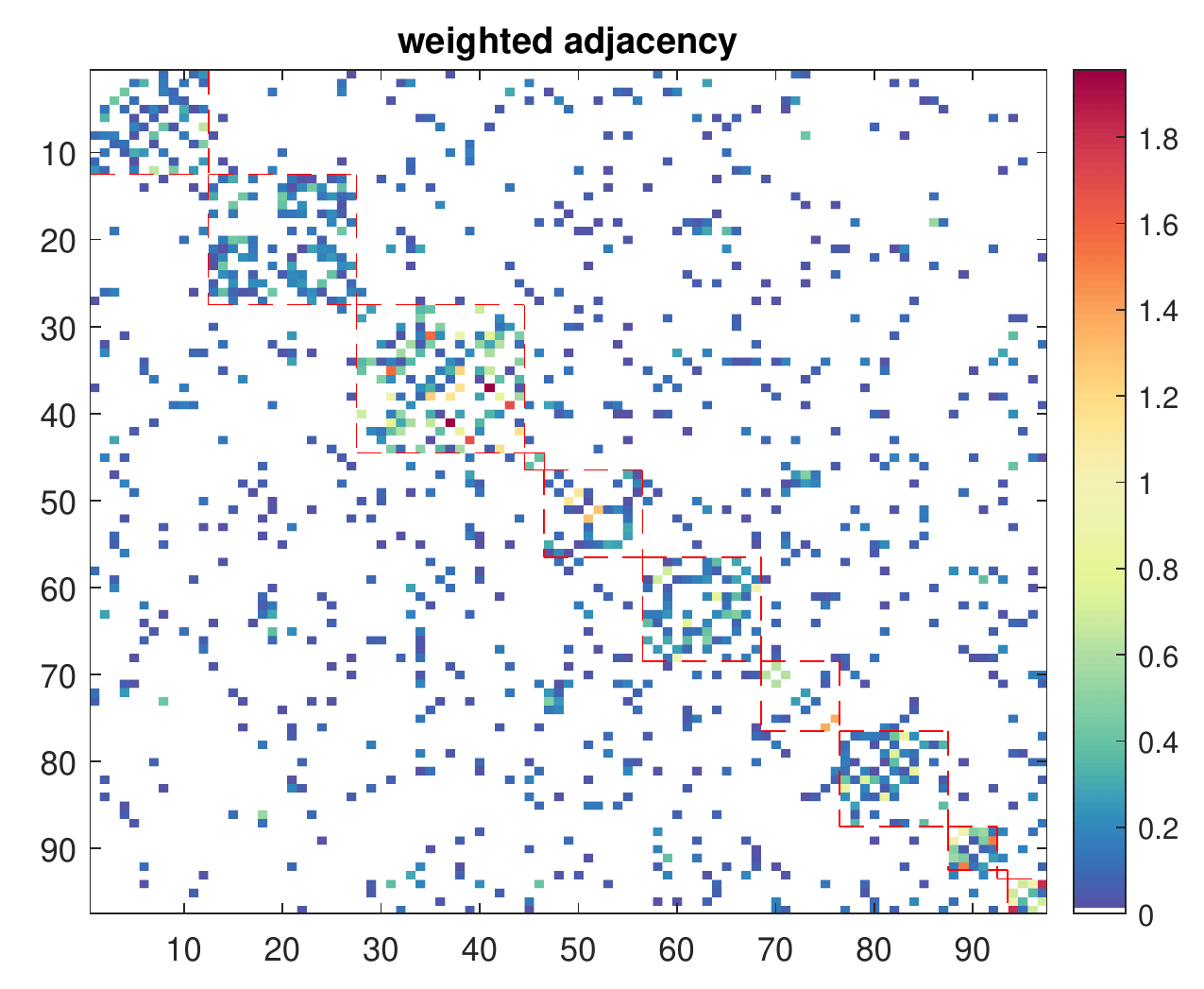}  
  \caption{Estimated $|\Omega_{ij}|$ as edge weight; 667 edges.}
  \label{fig5b}
\end{subfigure}
\caption{(b) Estimated weighted adjacency matrix ${\bm W}=-{\bm \Omega}^-$ for financial time series using proposed Const.\ Adap.\ Lasso (CAL) approach with BIC for $\lambda$ selection. The red squares (in dashed lines) show the 11 sectors -- they are not part of the adjacency matrix.}
\label{fig5}
\end{figure*}

We consider daily share prices (at close of the day) of 97 stocks in S\&P 100 index from Jan. 1, 2013 through Jan.\ 1, 2018, yielding 1259 samples. This data was gathered from Yahoo Finance website. If $y_m(t)$ is share price of $m$th stock on day $t$, we consider (as is conventional in such studies) $x_m(t) = \ln (y_m(t)/y_m(t-1))$ as the time series to analyze, yielding $n=1258$ and $p=97$. These 97 stocks are classified into 11 sectors (according to the Global Industry Classification Standard) and we order the nodes to group them as information technology (nodes 1-12), health care (13-27), financials (28-44), real estate (45-46), consumer discretionary (47-56), industrials (57-68), communication services (69-76), consumer staples (77-87), energy (88-92), materials (93), utilities (94-97). For each $m$, $x_m(t)$ was centered and normalized to unit variance. First we applied proposed CL and CAL approaches as well as LCGM with MCP \cite{Ying2020} to the data for varying penalty parameter $\lambda$ to evaluate number of detected edges. The results are shown in Fig.\ \ref{fig5a}. While the edge count decreases with increasing $\lambda$ for AL and CAL, that for LCGM increases with $\lambda$ for large $\lambda$'s, a totally unexpected and anomalous behavior similar to that encountered in \cite{Ying2020, Zhang2020} for CGL (or any Laplacian-constrained approach with convex penalty). We suspect the underlying  graphical model is not really a combinatorial Laplacian. Therefore, we only implemented our proposed CAL approach, and selected $\lambda$ using BIC (as for synthetic data), except now we take $\lambda_u = \lambda_{sm}/4$ and $\lambda_\ell = \lambda_u/40$). Fig.\ \ref{fig5b} shows estimated $|\Omega_{ij}^-|$ (=$W_{ij}$) with 667 edges. While the ground truth is unknown, the weighted adjacency matrix exhibits a modular structure that seems to conform to the sector classification according to the Global Industry Classification Standard.

\section{Conclusions} \label{concl}
The problem of learning a sparse undirected graph under graph Laplacian-related constraints on the sparse precision matrix was considered. Under these constraints the off-diagonal elements of the precision matrix are non-positive  and the precision matrix may not be full-rank. We investigated  modifications to widely used penalized log-likelihood approaches to enforce total positivity but not the Laplacian structure. The graph Laplacian can then be extracted from the off-diagonal precision matrix. An ADMM algorithm was presented for constrained optimization under Laplacian-related constraints and LSP penalty. Numerical results based on synthetic data show that the proposed constrained adaptive lasso approach significantly outperforms existing Laplacian-based approaches. We also evaluated our approach on real financial data. Our approach is applicable independent of the prior knowledge of the nature of the graph Laplacian (how many components, generalized or not), as illustrated by our synthetic data results based on one- and two- component Laplacian precision matrices (with one and two zero eigenvalues, respectively). However our theoretical results hold only under the assumption that ${\bm \Omega} \succ {\bm 0}$

\appendix

\section{Proof of Theorem 1} \label{proof}
Lemma 1 follows from \cite[Lemma 1]{Ravikumar2011}. \\
{\it Lemma 1}: Under Assumption (A2), the sample covariance $\hat{\bm{\Sigma}}$ satisfies the tail bound
\begin{equation}  \label{naeq58bx}
   P \left(  \max_{i,j} \Big| [ \hat{\bm{\Sigma}} - \bm{\Sigma}_0 ]_{kl} \Big| 
	    > C_0 \sqrt{\frac{\ln( p_n)}{n}} \right) \le \frac{1}{p_n^{\tau -2}} 
\end{equation}
for $\tau > 2$, if the sample size  $n >  N_1$, where $C_0$ is defined in (\ref{naeq58}) and $N_1$ is defined in (\ref{neq15ab2}). $\quad \bullet$

We now turn to the proof of Theorem 1. \\
{\it Proof of Theorem 1}. Let $\bm{\Omega} = \bm{\Omega}_0 + \bm{\Delta}$ with both $\bm{\Omega}, \, \bm{\Omega}_0 \succ {\bm 0}$ and both in ${\cal V}_p$, and $Q(\bm{\Omega}) := f_{LSP}({\bm \Omega}) - f_{LSP}({\bm \Omega}_0)$.
The estimate $\hat{\bm{\Omega}}_{\lambda}$, denoted by $\hat{\bm{\Omega}}$ hereafter suppressing dependence upon $\lambda$, minimizes $Q(\bm{\Omega})$, or equivalently, $\hat{\bm{\Delta}} = \hat{\bm{\Omega}} - \bm{\Omega}_0$ minimizes $G(\bm{\Delta}) := Q(\bm{\Omega}_0 + \bm{\Delta})$.
We will follow, for the most part, the method of proof of \cite[Theorem 1]{Rothman2008} pertaining to lasso penalty. Consider the set
\begin{equation}  \label{naeq1001}
  \Theta_n(M) :=  \left\{ \bm{\Delta} \, :\, \bm{\Delta} = \bm{\Delta}^\top, \; \|\bm{\Delta} \|_F = M r_n \right\}
\end{equation}
where $M$ and $r_n$ are as in (\ref{neq15ab0}) and (\ref{neq15ab1}), respectively.  Since $G(\hat{\bm{\Delta}}) \le G(\bm{0}) = 0 $, if we can show that $\inf_{\bm{\Delta}}  \{ G(\bm{\Delta}) \, :\, \bm{\Delta} \in \Theta_n(M) \} > 0$, then the minimizer $\hat{\bm{\Delta}}$ must be inside $\Theta_n(M)$, and hence $\| \hat{\bm{\Delta}} \|_F \le   M r_n$.
It is shown in \cite[(9)]{Rothman2008} that $\ln (|\bm{\Omega_0 + \Delta}|) - \ln (|\bm{\Omega}_0|) = \mbox{tr} (\bm{\Sigma_0  \Delta}) - A_1$ 
where, with $\bm{H}(\bm{\Omega}_0, \bm{\Delta}, v ) = (\bm{\Omega_0}+v \bm{\Delta})^{-1} \otimes (\bm{\Omega_0}+v \bm{\Delta})^{-1}$ and $v$ denoting a scalar,
\begin{align} \label{naeq1110}
     A_1 := & \mbox{vec}(\bm{\Delta})^\top \left( \int_0^1 (1-v) 
			  \bm{H}(\bm{\Omega}_0, \bm{\Delta}, v ) \, dv \right)  \mbox{vec}(\bm{\Delta}) \, .
\end{align}
Noting that $\bm{\Omega}^{-1} = \bm{\Sigma}$ and using 
\[
                 P_\lambda(\theta) = \lambda \ln(1+|\theta|/\epsilon) \, ,
\] 
we can rewrite $G({\bm{\Delta}})$ as 
\begin{align}
    & G({\bm{\Delta}})  = \sum_{i=1}^3 A_i \, ,  \quad 
		   A_2 :=\mbox{tr} \left( (\hat{\bm{\Sigma}} - \bm{\Sigma}_0 ) \bm{\Delta}  \right) \label{mainG}  \\
   &  A_3  := \sum_{i \ne j}^{p_n} (P_{{\lambda}_n} ( { \Omega}_{0ij} + { \Delta}_{ij} ) 
		         - P_{{\lambda}_n} ( { \Omega}_{0ij} )) \, . \label{naeq1120} 
\end{align}
Following \cite[p.\ 502]{Rothman2008}, we have
\begin{equation}
     A_1  \ge \frac{ \| \bm{\Delta} \|_F^2 }{  2 (\| \bm{\Omega}_0 \| +  \| \bm{\Delta} \|)^2 }
		  \ge \frac{ \| \bm{\Delta} \|_F^2 }{  2 \left( \beta_{\min}^{-1} + M r_n \right)^2 } \label{boundA1}
\end{equation}
where we have used the fact that $\| \bm{\Omega}_0 \| = \| \bm{\Sigma}_0^{-1} \| = \phi_{\max }(\bm{\Sigma}_0^{-1}) = (\phi_{\min }(\bm{\Sigma}_0))^{-1} \le \beta_{\min}^{-1} $ and $\| \bm{\Delta} \| \le \| \bm{\Delta} \|_F = M r_n = {\cal O}(r_n)$.
We now consider $A_2$ in (\ref{mainG}). We have
\begin{align}
     A_2  & =  L_{21} + L_{22} , \;\; L_{22} = \sum_{\{i,j\} \in {{\cal E}}^c_0 } 
			   [\hat{\bm{\Sigma}} - \bm{\Sigma}_0]_{ij} \Delta_{ji}  \, , \\
		L_{21} & = \sum_{\{i,j\} \in {\cal E}_0 } 
			   [\hat{\bm{\Sigma}} - \bm{\Sigma}_0]_{ij} \Delta_{ji} 
			  + \sum_{i} [\hat{\bm{\Sigma}} - \bm{\Sigma}_0]_{ii} \Delta_{ii}  \, ,
\end{align}
where ${{\cal E}}^c_0$ denotes the complement of set ${{\cal E}}_0$
For an index set ${\bm B}$ and a matrix ${\bm C} \in \mathbb{R}^{p_n \times p_n}$, we write ${\bm C}_{\bm B}$ to denote a matrix in $\mathbb{R}^{p_n \times p_n}$ such that $[{\bm C}_{\bm B}]_{ij} = C_{ij}$ if $(i,j) \in {\bm B}$, and $[{\bm C}_{\bm B}]_{ij}=0$ if $(i,j) \not\in {\bm B}$. Using this notation, 
to bound $L_{21}$, using Cauchy-Schwartz inequality and Lemma 1, with probability $> 1- 1/p_n^{\tau-2}$,
\begin{align}
    | L_{21} |  \le &  \| \bm{\Delta}_{{\cal E}_0}^- + \bm{\Delta}^+ \|_1 \, 
	  \max_{i,j} \big|  [\hat{\bm{\Sigma}} - \bm{\Sigma}_0]_{ij}  \big|   \nonumber \\
		\le & \sqrt{s_{n0} + p_n}  \, \|\bm{\Delta}\|_F  C_0 \sqrt{ \ln(p_n)/n}
				= C_0 \|\bm{\Delta}\|_F r_n \, . \label{naeq1205}
\end{align}
We consider $L_{22}$ later as a part of $A_3$ where
\begin{align}
     A_3 & =  L_{31} + L_{32}, \quad  L_{32}  = \sum_{\{i,j\} \in {{\cal E}}^c_0 } 
			   P_{{\lambda}_n} (  { \Delta}_{ij} ) , \\
      L_{31} & = \sum_{\{i,j\} \in {\cal E}_0 } 
			   (P_{{\lambda}_n} ( { \Omega}_{0ij} + { \Delta}_{ij} ) 
		         - P_{{\lambda}_n} ( { \Omega}_{0ij} ))  
\end{align}
where we have used that fact that, for $\{i,j\} \in {{\cal E}}^c_0$, $\Omega_{0ij} = 0$, hence, $P_{{\lambda}_n} ( { \Omega}_{0ij} )=0$.

Since $\ln(1+x) \ge x/(1+x)$ for $x >-1$, we have $\ln(1+x) \ge x/C_1$ for $0 \le x \le C_1-1$ for any $C_1 > 1$. Therefore,
\begin{align*}
  P_{{\lambda}_n} (  { \Delta}_{ij} ) & \ge \frac{ \lambda_n |\Delta_{ij}|}{\epsilon C_1} \mbox{ for } 
	   \frac{  |\Delta_{ij}|}{\epsilon } \le C_1 -1 \, .
\end{align*}
Notice that $|\Delta_{ij}| \le \|{\bm \Delta}\|_F$, $\|{\bm \Delta}\|_F = Mr_n$ on $\Theta_n(M)$, and for $n \ge N_2$, $M r_n /\epsilon \le C_1-1$, by (\ref{neq15ab2}). Therefore, for $n \ge N_2$, 
\begin{align*}
  P_{{\lambda}_n} (  { \Delta}_{ij} ) & \ge \frac{ \lambda_n |\Delta_{ij}|}{\epsilon C_1} \mbox{ for } n \ge N_2 \, .
\end{align*}
Consider $L_{32}$ with $ L_{22}$
\begin{align}
     L_{32}  -  | L_{22} | & \ge \sum_{\{i,j\} \in {{\cal E}}^c_0 }  
		  \Big( \frac{ \lambda_n |\Delta_{ij}|}{\epsilon C_1}
		  - | [\hat{\bm{\Sigma}} - \bm{\Sigma}_0]_{ij} | \, | { \Delta}_{ij} | \Big) \nonumber \\
		&	\ge  \Big( \frac{ \lambda_n }{\epsilon C_1}- C_0 \sqrt{\frac{\ln(p_n)}{n}} \Big) 
		  \sum_{\{i,j\} \in {{\cal E}}^c_0 } | { \Delta}_{ij} | 
		 \ge 0 	\label{eqdif1}
\end{align}
with probability $> 1- 1/p_n^{\tau-2}$, since $C_0 C_1 \sqrt{ \ln( p_n) /n } < (\lambda_n/\epsilon)$ by (\ref{neq15abc}).

Now we bound $|L_{31}|$. A Taylor series expansion of $P_{\lambda}(\theta)$ for $\theta > 0$, around $\theta_0 > 0$, is given by 
$P_{\lambda}(\theta) = P_{\lambda}(\theta_0) + P_{\lambda}^{\prime}(\theta_0) (\theta - \theta_0) 
	        + P_{\lambda}^{\prime\prime}(\tilde{\theta}) \frac{(\theta - \theta_0)^2}{2} $
where $\tilde{\theta} = \theta_0 + \gamma (\theta - \theta_0)$ for some $\gamma \in [0,1]$. Setting $\lambda = {\lambda}_n$, $\theta_0 = |\Omega_{0ij}|$ and  $\theta = |\Omega_{0ij}+\Delta_{ij}| $, and noting that $P_{\lambda}^{\prime\prime}({\theta}) = -\lambda/(|{\theta}|+\epsilon)^2 \, < 0$ for any ${\theta} >0$ and $P_{\lambda}^{\prime}({\theta}) = \lambda/(|{\theta}|+\epsilon) > 0$, we have
\begin{align*}
  P_{{\lambda}_n} (\Omega_{0ij}+\Delta_{ij} ) \le &  P_{{\lambda}_n} (\Omega_{0ij}) \\
	  & + P_{{\lambda}_n}^{\prime}(\Omega_{0ij}) (|\Omega_{0ij}+\Delta_{ij} | - |\Omega_{0ij}|) \, .
\end{align*}
Thus we have
\begin{align}
   L_{31}  = &  \sum_{ \{i,j\} \in {\cal E}_0 } \frac{\lambda_n}{|\Omega_{0ij}|+\epsilon} 
	  ( |\Omega_{0ij}+\Delta_{ij} | - |\Omega_{0ij}| )  \nonumber \\ \\
		 \le & \sum_{ \{i,j\} \in {\cal E}_0 } \frac{\lambda_n}{|\Omega_{0ij}|+\epsilon} |\Delta_{ij} | 
		    \nonumber \\
	\Rightarrow |L_{31}| \le & 	   \frac{\lambda_n}{\epsilon} \sum_{\{i,j\} \in {\cal E}_0 } |\Delta_{ij} |
	 \le \frac{\lambda_n}{\epsilon}  \sqrt{s_{n0}} \, \|{\bm \Delta} \|_F \, . \label{eqdif2}
\end{align}

Combining $A_2$ and $A_3$, and using $L_{32}  -  | L_{22} | \ge 0$, we have
\begin{align}
  A_2+  A_3 = & L_{21}+L_{22}+L_{31} + L_{32} \nonumber \\
	  \ge &  -|L_{21}| -|L_{22}| - |L_{31}| + L_{32}   \nonumber \\
		 \ge & -|L_{21}|  - |L_{31}| \nonumber \\
		\ge &
		       -C_0 \|\bm{\Delta}\|_F r_n  - \frac{\lambda_n}{\epsilon} \sqrt{s_{n0}} \, \|{\bm \Delta} \|_F  \, .  \label{naeq5330}
\end{align}
By (\ref{neq15abc}), 
\[
  \frac{\lambda_n}{\epsilon}  \le C_1 C_0  \frac{r_n }{\sqrt{s_{n0}} }  \Rightarrow 
	\frac{\lambda_n}{\epsilon} \sqrt{s_{n0}} \, \|{\bm \Delta} \|_F \le C_1 C_0  r_n  \|{\bm \Delta} \|_F \, .
\]
Using (\ref{mainG}), the bound (\ref{boundA1}) on $A_1$ and (\ref{naeq5330}) on $A_2+  A_3$, and $\| \bm{\Delta} \|_F = M r_n$, we have with probability $> 1- 1/p_n^{\tau-2}$, 
\begin{align}
     G({\bm{\Delta}}) \ge &  \; \| \bm{\Delta} \|_F^2 \left[ \frac{1}{2 (\beta_{\min}^{-1} + M r_n )^2}  
		     - \frac{C_2C_0}{M} \right]  \, .\label{naeq1370} 
\end{align}
For $n \ge N_2$, if we pick $M$ as specified in (\ref{neq15ab0}), we obtain $M r_n \le M r_{N_2} \le \delta_1/\beta_{\min}$. Then
\begin{align*}
    \frac{1}{2 (\beta_{\min}^{-1} + M r_n )^2} \ge & \frac{\beta_{\min}^2}{2 (1+\delta_1)^2} \\
		    = & \frac{(2 C_2 + \delta_2)C_0}{2 M} >  \frac{C_2 C_0}{M} \, ,
\end{align*}
implying $G({\bm{\Delta}})  > 0$. This completes the proof.    $\quad \blacksquare$

\section{Proof of Theorem 2} \label{proof2}
Consider the $(i,j)$th element ${\hat{\Omega}}_{\lambda ij}$ of the LSP estimate $\hat{\bm{\Omega}}_\lambda$. Since $\hat{\bm{\Omega}}_\lambda$ minimizes the cost $f_{LSP}({\bm \Omega})$ given by (\ref{neq2000a}) under the constraint ${{\Omega}}_{ij} \le 0$, for ${\hat{\Omega}_{\lambda ij}} \ne 0$ (i.e., ${\hat{\Omega}_{\lambda ij}} < 0$), we must have
\begin{align}
   0 & = \frac{\partial f_{LSP}({\bm \Omega})}{\partial \hat{\Omega}_{\lambda ij}} 
	   = \hat{\Sigma}_{ji} - [ \hat{\bm \Omega}_{\lambda}^{- 1} ]_{ji} 
		    + P_{\lambda_n}^{\prime} (|\hat{\Omega}_{\lambda ij}|)  \nonumber \\
		& = \hat{\Sigma}_{ij} - \check{\Sigma}_{\lambda ij} 
		  + \frac{\lambda_n}{|\hat{\Omega}_{\lambda ij}|+\epsilon} \;
			 \frac{\hat{\Omega}_{\lambda ij}}{|\hat{\Omega}_{\lambda ij}|} 
				=: A
     \label{naeq2000}   
\end{align}
where 
\[
      \check{\bm \Sigma}_{\lambda}  := \hat{\bm \Omega}_{\lambda}^{-1}
\] 
and we use the notation
\[
  \frac{\partial f_{LSP}({\bm \Omega})}{\partial \hat{\Omega}_{ij}} =
		\frac{\partial f_{LSP}({\bm \Omega})}{\partial {\Omega}_{ij}} 
		       \Big|_{{\bm \Omega} = \hat{\bm \Omega}_{\lambda}} \, .
\]

To prove the desired result, the term $ \lambda_n (\hat{\Omega}_{\lambda ij}/((|\hat{\Omega}_{\lambda ij}|+\epsilon)|\hat{\Omega}_{\lambda ij}|)$ on the right-side of (\ref{naeq2000}) must dominate the term $\hat{\Sigma}_{ij} - \check{\Sigma}_{\lambda ij}$ whenever true value $\Omega_{0ij} = 0$. Then the sign of $\frac{\partial f_{LSP}({\bm \Omega})}{\partial \hat{\Omega}_{\lambda ij}}$ in (\ref{naeq2000}) is the same as $\mbox{sign}(\hat{\Omega}_{\lambda ij})$ with probability tending to one, which yields the desired result, as is shown in what follows. At the optimal solution ${\hat{\Omega}_{\lambda ij}} < 0$, by the KKT conditions, one must have $A$ in (\ref{naeq2000}) equal to zero. Suppose that for $\{i,j\} \in {\cal E}_0^c$, one has $\hat{\Omega}_{ij} < 0$ when $A = 0$. This implies that for some $\delta > 0$, $\hat{\Omega}_{\lambda ij } + \delta < 0$, since, by Theorem 1, $\hat{\Omega}_{\lambda ij }$ converges to ${\Omega}_{0 ij } = 0$ for $\{i,j\} \in {\cal E}_0^c$. Since $\hat{\Omega}_{\lambda ij }$ minimizes $f_{LSP}({\bm \Omega})$, and $\frac{\partial f_{LSP}({\bm \Omega})}{\partial \hat{\Omega}_{ij}} = 0$ for $\hat{\Omega}_{\lambda ij } < 0$, we must have $I := \frac{\partial f_{LSP}({\bm \Omega})}{\partial (\hat{\Omega}_{ij}+ \delta)} > 0$ for $\delta > 0$. If $\lambda_n$ dominates $\hat{\Sigma}_{ij} - \check{\Sigma}_{\lambda ij}$ in (\ref{naeq2000}), $I > 0$ implies that $\hat{\Omega}_{\lambda ij} + \delta > 0$, contradicting the assumption that $\hat{\Omega}_{\lambda ij } + \delta < 0$. Therefore, $\hat{\Omega}_{\lambda ij } \nless 0$, hence, $\hat{\Omega}_{\lambda ij } = 0$ for $\{i,j\} \in {\cal E}_0^c$, with probability tending to one. 

It remains to investigate the conditions under which $\lambda_n$ dominates $\hat{\Sigma}_{ij} - \check{\Sigma}_{\lambda ij}$ independent of $i$ and $j$.
Rewrite 
\begin{equation}
   \hat{\Sigma}_{ij} - \check{\Sigma}_{\lambda ij} =
	 \underbrace{\hat{\Sigma}_{ij} - {\Sigma}_{0 ij}}_{=: I_2} 
	   + \underbrace{{\Sigma}_{0ij} - \check{\Sigma}_{\lambda ij}}_{=: I_3} \, .
\end{equation}
By Lemma 1, $\max_{i,j} |I_2| = {\cal O}_P \left(\sqrt{\frac{\ln(p_n)}{n}} \right)$.
By \cite[Lemma 1]{Lam2009}, 
\begin{align}
   |I_3| \le & \|\bm{\Sigma}_0 - \check{\bm{\Sigma}}_{\lambda} \|
	         = \|\check{\bm{\Sigma}}_{\lambda}  ( \hat{\bm \Omega}_{\lambda} -
					      \bm{\Omega}_0 ) \bm{\Sigma}_0\|  \nonumber \\
				\le & \|\check{\bm{\Sigma}}_{\lambda} \| \cdot \|  ( \hat{\bm \Omega}_{\lambda} -
					      \bm{\Omega}_0 ) \| \cdot \|\bm{\Sigma}_0\|	\, .\label{naeq2100}   
\end{align}
By Assumption (A2), $\|\bm{\Sigma}_0\| = {\cal O}(1)$. Furthermore,
\begin{align}
   \|\check{\bm{\Sigma}}_{\lambda} \| = & \|\hat{\bm \Omega}_{\lambda}^{-1} \|
	   = \phi_{\min}^{-1}(\hat{\bm \Omega}_{\lambda}) \nonumber \\ 
		\le &  \left( \phi_{\min} (\bm{\Omega}_0) + \phi_{\min} 
		  (\hat{\bm \Omega}_{\lambda}-\bm{\Omega}_0) \right)^{-1}  \nonumber \\
			=  & ( {\cal O}_P(1) + {\cal O}_P(\eta_n))^{-1} = {\cal O}_P(1)
	\, , \label{naeq2110}   
\end{align}
where we have used the fact that since $\| \hat{\bm \Omega}_{\lambda}-\bm{\Omega}_0 \| = {\cal O}_P(\eta_n)$, $\phi_{\min} (\hat{\bm \Omega}_{\lambda}-\bm{\Omega}_0) \le \| \hat{\bm \Omega}_{\lambda}-\bm{\Omega}_0 \| = {\cal O}_P(\eta_n)$, and by Weyl's inequality, $\phi_{\min}({\bm A}+{\bm B}) \ge \phi_{\min}({\bm A}) + \phi_{\min}({\bm B})$. Hence,
\begin{equation}
  \max_{i,j} |I_3| = {\cal O}_P \left( \| \hat{\bm \Omega}_{\lambda}-\bm{\Omega}_0 \| \right)
	            = {\cal O}_P \left(\eta_n \right) \, .
\end{equation}
It then follows that
\begin{equation}
  | \hat{\Sigma}_{ij} - \check{\Sigma}_{\lambda ij} | \le |I_2| + |I_3|
	            = {\cal O}_P \left(\sqrt{\frac{\ln(p_n)}{n}} + \eta_n \right) \, .  \label{naeq2150}
\end{equation}
Suppose ${\cal O}(\lambda_n) = \sqrt{\ln(p_n)/n} + \eta_n $. Then 
$ \lambda_n (\hat{\Omega}_{\lambda ij}/((|\hat{\Omega}_{\lambda ij}|+\epsilon)|\hat{\Omega}_{\lambda ij}|)$ dominates $| \hat{\Sigma}_{ij} - \check{\Sigma}_{\lambda ij} |$ with probability tending to one. This completes the proof. $\quad \blacksquare$

\bibliographystyle{unsrt}

\end{document}